\documentclass{article}

\PassOptionsToPackage{numbers, compress}{natbib}

\usepackage[preprint]{neurips_2021}

\usepackage[utf8]{inputenc} 
\usepackage[T1]{fontenc}    
\usepackage{hyperref}       
\usepackage{url}            
\usepackage{booktabs}       
\usepackage{amsfonts}       
\usepackage{nicefrac}       
\usepackage{microtype}      
\usepackage{xcolor}         
\usepackage{multirow}
\usepackage{subfigure}
\usepackage{caption}
\usepackage{figsize}
\usepackage{wrapfig}
\usepackage{makecell}
\usepackage{color, colortbl}
\usepackage{amsmath}
\usepackage{enumitem}

\title{Multi-View Node Pruning for\\ Accurate Graph Representation}

\author{%
   Hanjin Kim* \And
   Jiseong Park* \And
   Seojin Kim \And
   Jueun Choi \And
   Doheon Lee \And
   Sung Ju Hwang \\
}

\begin{document}

\maketitle
\begin{abstract}
Graph pooling, which compresses a whole graph into a smaller coarsened graph, is an essential component of graph representation learning. To efficiently compress a given graph,  graph pooling methods often drop their nodes with attention-based scoring with the task loss. However, this often results in simply removing nodes with lower degrees without consideration of their feature-level relevance to the given task. 
To fix this problem, we propose a ~\textit{Multi-View Pruning} (MVP), a graph pruning method based on a multi-view framework and reconstruction loss. Given a graph, MVP first constructs multiple graphs for different views either by utilizing the predefined modalities or by randomly partitioning the input features, to consider the importance of each node in diverse perspectives. Then, it learns the score for each node by considering both the reconstruction and the task loss. MVP can be incorporated with any hierarchical pooling framework to score the nodes. We validate \textit{MVP} on multiple benchmark datasets by coupling it with two graph pooling methods, and show that it significantly improves the performance of the base graph pooling method, outperforming all baselines. Further analysis shows that both the encoding of multiple views and the consideration of reconstruction loss are the key to the success of MVP, and that it indeed identifies nodes that are less important according to domain knowledge.
\end{abstract}

\section{Introduction}

Graph Neural Networks (GNNs) ~\cite{scarselli2008graph,li2015gated, kipf2016semi, NIPS2017_5dd9db5e, gilmer2017neural, velivckovic2017graph, xu2018how}, which enable learning deep representations for graph-structured data (e.g. molecular graphs, social networks, and knowledge graphs), have received considerable attention recently due to the increasing interest in graph-based applications (e.g. graph classification, link prediction, and graph generation). While earlier works on GNNs focused on how to better represent each node by aggregating the information from its neighbors, representing the graph as a whole is also an important problem. To tackle this problem, researchers have proposed various \emph{graph pooling} methods that allow compressing a given graph into a smaller graph or even a single vector~\cite{bianchi2020spectral}. Existing graph pooling methods compress the given set of node representations either by clustering~\cite{ying2018hierarchical, baek2021accurate}, dropping nodes~\cite{gao2019graph,lee2019self}, or using self-attentions~\cite{lee2019self}. 

\begin{figure}[bt!]
    \centering
    \subfigure[
    \small AA]{\includegraphics[height=.24\columnwidth, trim = 0cm 0cm 33cm 0cm, clip]{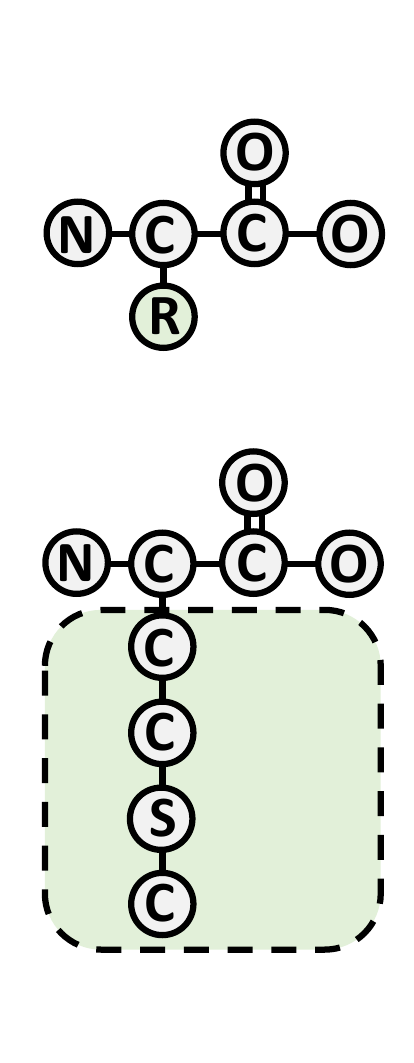}}
    \subfigure[\small Reconstruction-based multi-view pruning score]{\includegraphics[height=.24\columnwidth,trim = 0.5cm 0cm 2.5cm 0cm, clip]{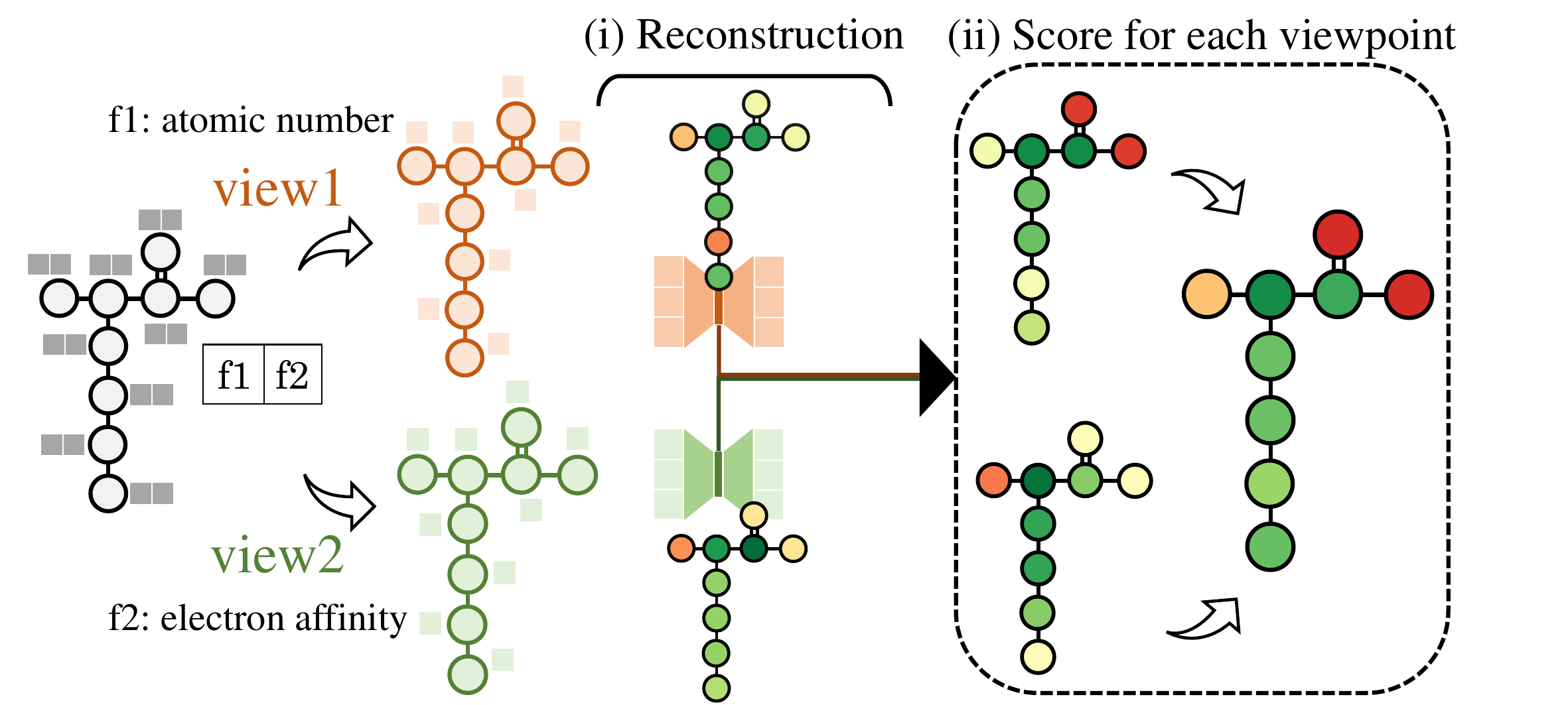}}
    \subfigure[\small Att-based]{\includegraphics[height=.24\columnwidth,trim = 0cm 0cm 30cm 0cm, clip]{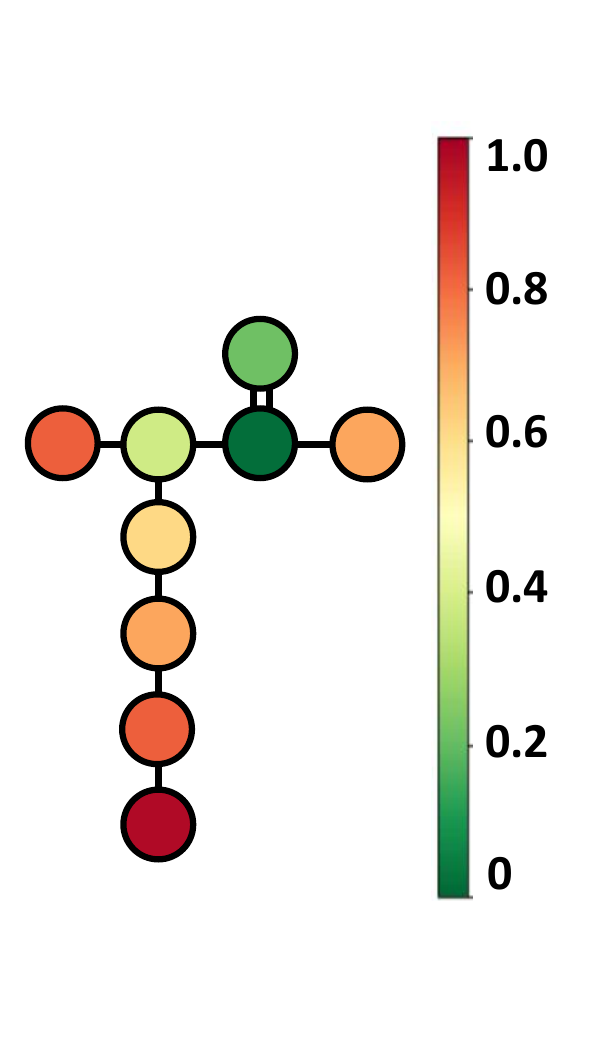}}
    \hspace{0.5em}
    \subfigure[\small Pruning 
    ratio]{\includegraphics[height=.23\columnwidth]{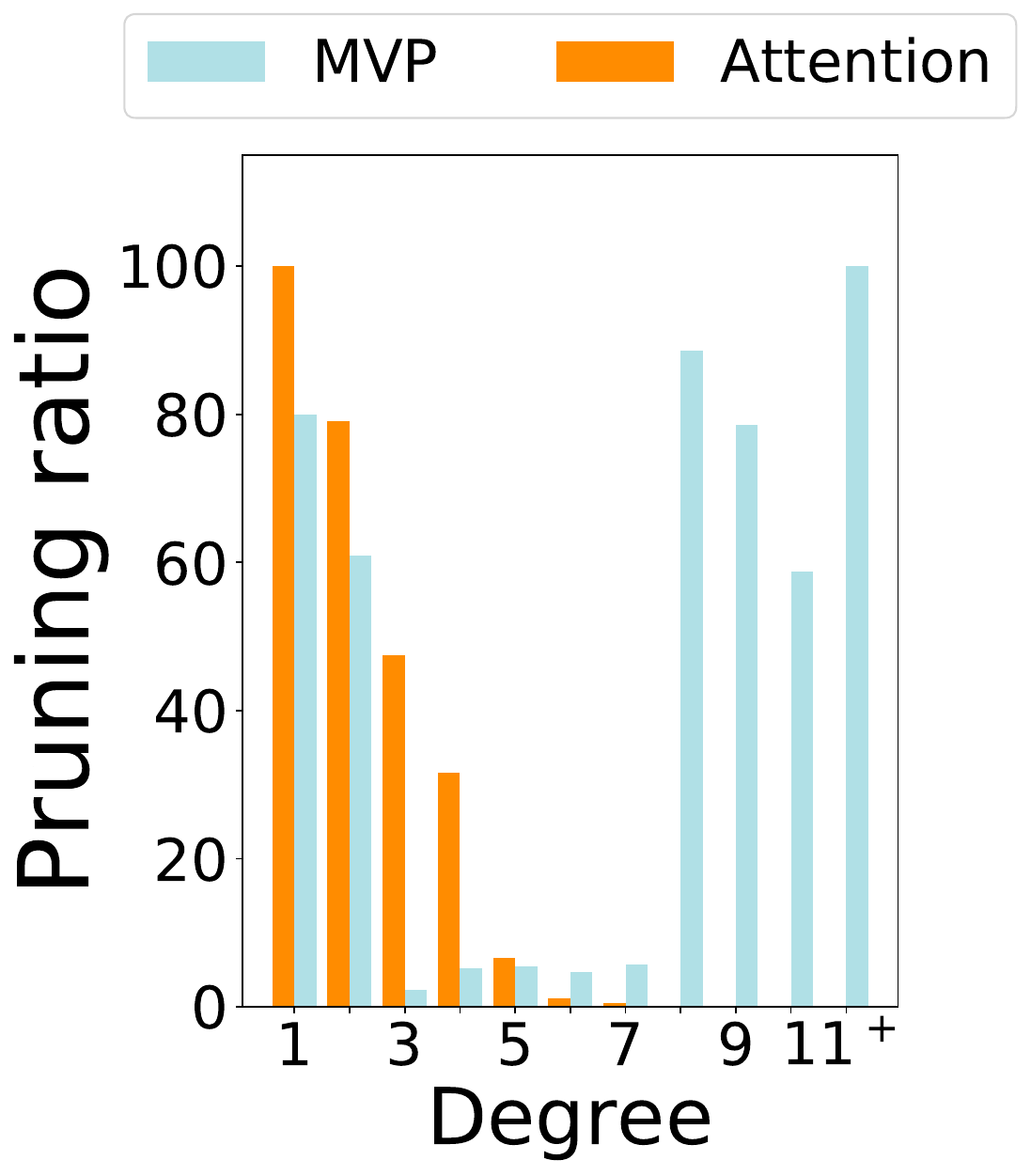}}
    \caption{\small \textbf{Concept.} (a) General structure of aminoacid (upper) and amino acid "Methionine" (lower). Green part is important to identify the characteristic of different amino-acid.(b) Reconstruction-based scores of nodes. (c) Attention-based scores of nodes. The scoring is conducted for amino acid classification task and the example score reflects the actual values. (d) The number of pruning nodes according to the performance of each scoring policy.}

    \vspace*{-0.7cm}
    \label{fig:mv_importance}
\end{figure}

A crucial challenge in graph pooling is that not all nodes are equally important for the given task, as some nodes are more informative while some others may be unnecessary or even harmful. For example, when classifying the given molecular graphs for toxicity prediction, only a certain substructure of the molecules could be responsible, and including any irrelevant atoms may only confuse the classifier. In such cases, it will be beneficial to remove less informative or distracting nodes from the given graph. 

Based on this motivation, we can \emph{prune} the nodes of a given graph to preserve the most informative ones. But how should we decide which nodes are more informative? Unlike pruning neurons in conventional neural networks that can be done by their average activation scores, there is no straightforward way to measure the importance of each node from a given graph. Existing pooling methods based on node pruning~\cite{gao2019graph, lee2019self} use attention-based scoring of nodes, that are obtained to minimize the task loss. Although this strategy sounds reasonable, we empirically observe that it often results in pruning nodes simply based on their degrees, since nodes with larger degrees contain information about more number of nodes. This is shown in Figure~\ref{fig:mv_importance}(c) and (d), which show the scoring obtained by an attention-based scoring, and pruning rates for nodes with different degrees, for classifying different amoniacids. However, this could be highly suboptimal since it ignores the rich node features that may be discriminative for the target tasks. For example, when classifying the molecular graph of a given aminoacid, it is important to examine the side-chain (green part of Figure \ref{fig:mv_importance}(a)) but the nodes of the side-chain are scored high due to their low degrees. 

Thus, we need a more careful scoring of the nodes, to better measure their importance. To this end, we propose to utilize \emph{graph reconstruction error} as a scoring measure. This allows us to detect nodes that are considered less informative in minimizing the task loss, since their reconstruction errors will be high. Since we consider both the reconstruction error for the node features and the adjacency matrix, this will consider the node features rather than simple degrees of each node.

However, when scoring the nodes to identify the anomalies, considering all features in a single pool may be problematic. 
Existing attention-based scoring preserves the nodes with high degrees, since they contain information about larger number of nodes due to GNN aggregation. However, this will only worsen the oversmoothing problem~\cite{zhao2019pairnorm, chen2020measuring}, and thus the information of the nodes with critical features may be highly diluted or even fade out, in the remaining nodes. MVP resolves the problem with reconstruction-based scoring with balanced consideration of all types of features via multi-view embedding. For example in the Figure \ref{fig:mv_importance}, in element perspective (view 1), the sulfur node (node 'S') is critical for classifying the graph while the rest are 
general structures. Thus the node 'S' is assigned low pruning score by our task-specific reconstruction-based scoring. In addition, since the electron affinity (view 2) of 'S' is also meaningful, the pruning score for node 'S' in this view is also low. Thus, the final score for node 'S' has a low score, which prevents it from being pruned away.
%

We validate the performance of our MVP on benchmark datasets for graph classification tasks.
The results show that our MVP significantly outperforms baseline pooling methods on most tasks, achieving new state-of-the-art performances. We further validate the quality of pruning by calculating the betweeness centrality and examining the location of the pruned nodes in each graph, whose results suggest that the discarded nodes are indeed less informative or superfluous. Further qualitative analysis suggests that our method prunes out nodes with less semantic importance.

In sum, our contributions can be summarized as follows: 
\begin{itemize}[leftmargin=1em]
\vspace*{-0.3cm}
\setlength\itemsep{-0.3em}
    \item We propose a novel graph pruning method for graph pooling, which utilizes a reconstruction-based node scoring to identify informative nodes for the given task. 
    \item While doing so, we propose to consider  multiple views of the graphs into account, by projecting a random subset of the input features to each view. 
    \item We empirically demonstrate that our multi-view pruning method with reconstruction-based scoring significantly improves base pooling methods on benchmark graph classification data.
    \item We further demonstrate that our multi-view pruning can capture more informative nodes than existing attention-based node pruning methods, both with quantitative and qualitative analysis.
\end{itemize}

\section{Related Work}
\vspace*{-0.2cm}
\paragraph{Graph Pooling in GNN} 
Graph neural networks (GNNs) learn representations for given graphs by exploiting their topological structure as well as the node and edge features. 
While earlier works are mostly concerned with learning the representation for each node~\cite{hamilton2017inductive, gilmer2017neural}, recent works on graph pooling focus on holistic graph representations.
The most simple approach to obtain graph-level representations, is to perform simple averaging or sum of the node representations. 
However, such simple approaches cannot capture higher-order interactions among the nodes, nor their importance to the given task. Thus, many pooling approaches utilizing clustering or node pruning to tackle this problem. 
Clustering-based approaches, such as
DiffPool~\cite{ying2018hierarchical}, MinCutPool~\cite{bianchi2020spectral} and GMT~\cite{baek2021accurate} coarsen the graph into a smaller graph or a single vector with spectral clustering or multi-head attention-based pooling.
Graph pruning approaches drop less important nodes by scoring them, usually via attentions, since not all nodes are equally important for the target task. Our MVP falls into the latter category, which we describe more in detail in the next paragraph.

\vspace*{-0.2cm}
\paragraph{Graph Pruning and Anomaly Detection} 
The primary goal of pruning is to remove unnecessary, superfluous elements, thereby optimizing the model's use of memory and computation. 
TopKPool~\cite{gao2019graph} is a graph pruning-based graph pooling method, which retains only the top-k nodes from the graph based on the importance of the nodes. 
SAGPool~\cite{lee2019self} improves upon the TopKPool by calculating the scores of the nodes based on both the node's score and its neighbor's scores, via graph convolution. 
However, both SAGPool and TopKPool are limited in that they rely on attention scores to decide which nodes to prune, which is dominated by the topological features (e.g. degree) and may miss out critical node features.
Anomaly detection in the graph domain aims to rank the anomalies of the nodes based on reconstruction errors, assuming that instances with large residuals during reconstruction are anomalous.
Hou et. al.~\cite{hou2020fast} conduct deep encoder architecture to detect outlier of the graph by learning low-rank latent subspace representation embedded from clustered heterogeneous nodes in different views. 
Peng et. al.~\cite{peng2020deep} divide the graph into multiple graphs with a different subset of the original features based on the view of the features, 
then concatenate all representations embedded from GNN to detect the most abnormal node in the graph. However, none of the existing works utilizes reconstruction errors for graph pooling, as our MVP does.

\vspace*{-0.2cm}
\paragraph{Multi-View Representation Learning} 
Multi-view representation learning aims to investigate the multi-view data embedding to a new-shared latent space for educing the capable representations~\cite{ngiam2011multimodal}. 
In general, this technique learns more comprehensively than single-view learning because data from different views usually includes information that complements each view. Also, it has become a promising topic since the expressiveness of data representation affects the performance of the machine learning~\cite{li2018survey}.
In recently, multi-view data becomes more available in real-world problems, so this mechanism is increasingly paid attention to. Multi-view representation learning adopted to various deep neural network models and they are usually outperforming the existing model
in applying various data, including complicated biological data~\cite{qi2016deep}.
When dealing with multi-view data, it is important to distinguish whether each data is necessary or unnecessary for each view. 
Therefore, we propose to detect an anomalous node in the multi-view latent space, by computing the node embedding separately for each view, and then concatenating them.

\section{Method}

\makeatletter
\newcommand{\thickhline}{%
    \noalign {\ifnum 0=`}\fi \hrule height 1pt
    \futurelet \reserved@a \@xhline
}

To formally define the problem, let $G = (V, E, F)$ denote a graph consisting of $|V|=n$ nodes and $|E|=m$ edges, with d-dimensional node features $F=\{f_1, f_2, \cdots, f_d\}$ which can be represented by $k$ distinct features for each view. A graph $G$ is characterized by its node features $\mathbf{X}\in \mathbb{R}^{n\times d}$ and the adjacency matrix $\mathbf{A}\in \mathbb{R}^{n\times n}$ which defines the edges across the nodes. Given a graph $G$ and its corresponding label $y$, the pooling function aims to learn a graph-level representation $h_G$, that can be used to predict the label of $G$, $y_G=g(h_G)$. This can be done by compressing the given graph into a smaller graph, or a vector, by dropping out unimportant nodes.  

To obtain only the most informative nodes from the given graph, we propose a graph pooling method based on graph pruning.
Our method first identifies superfluous nodes based on the pruning score, and uses it as a pruning criterion to drop each node. The pruning score of each node is obtained as the task-specific reconstruction errors based on the graph representation obtained through multi-view embedding. We can then combine the proposed graph pruning procedure with any graph pooling method, to enhance its performance. We will describe each procedure in detail in the following subsections.

\begin{figure}[bt!]
    \centering
     \includegraphics[width=\linewidth]{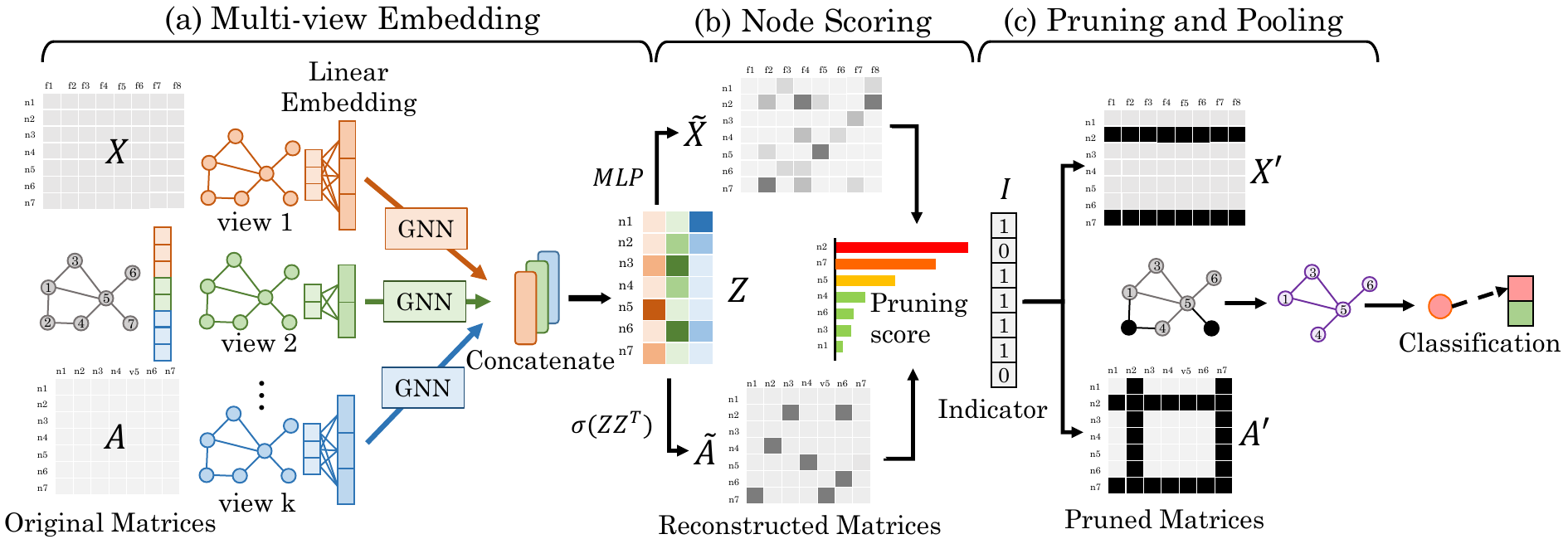}
    \vspace*{-0.5cm}
    \caption{\small \textbf{Overview of MVP}. First, (a) the input graph passes graph convolution layer for each view, and then (b) using the embedded feature matrix, concatenate them to make latent space $\mathbf{Z}$ and make indicator $\mathbf{I}$ by scoring nodes for pruning using $\mathbf{Z}$. Finally, (c) the model prunes original feature matrix and adjacency matrix for the input graph with the indicator $\mathbf{I}$. The pruned graph then passes the pooling layer so that it is represented as a vector which is used for graph classification.}
    \label{fig:schema}
    \vspace*{-0.55cm}
\end{figure}

\vspace*{-0.2cm}
\subsection{Multi-view feature extraction}
\begin{wrapfigure}{r}{.28\textwidth}
    \vspace*{-1.3cm}
    \frame{\includegraphics[width=.28\textwidth]{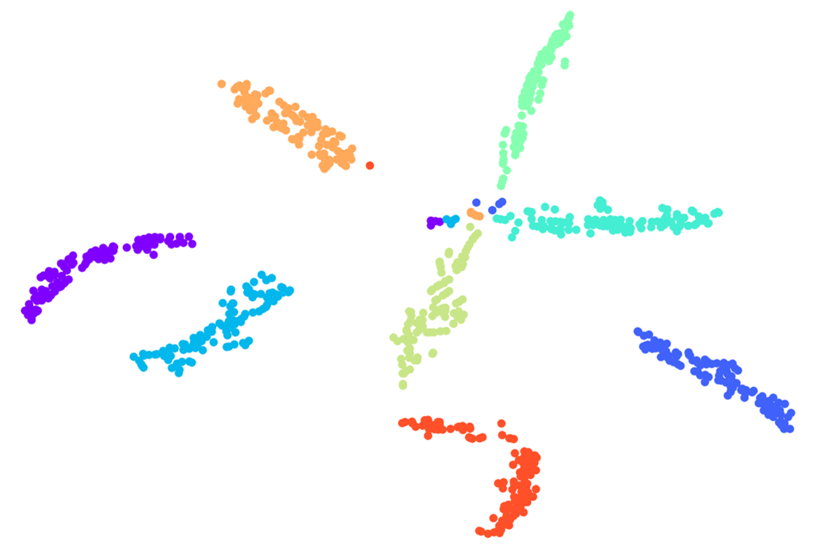}}
    \caption{\small t-SNE visualization of the embedded features for 8 views. Each color denotes a different view.}
    \label{fig:tsne_}
    \vspace*{-0.5cm}
\end{wrapfigure}
\vspace*{-0.2cm}
We first construct multiple views of the graph, to consider the multi-modality of the given data. Real-world graphs may come with intrinsic multi-modalities, in which case we can simply construct multiple views with each modality. However, in many cases, it may be difficult to find a separable modality due to ambiguity. In such a case, we can generate multiple views of a graph by randomly assigning the features to each modality. Moreover, when doing so, the features could be highly heterogeneous, as they could be either continuous, binary, or categorical. To resolve this issue, we linearly embedded each set of features using a single-layer neural network. The visualization of the t-distributed stochastic neighbor embedding (t-SNE)~\cite{maaten2008visualizing} shows that this approach can embed features from each view onto a distinct modality (Figure \ref{fig:tsne_}. Also see Section \ref{supple:tsne} of the supplementary materials).

The graph for each view is embedded through an independent GNN, to ensure extracting features that best reflect each modality. Each of these GNN consists of a single graph convolution layer which calculates the embedded feature matrix with the input feature matrix and the adjacency matrix with the equation $\mathbf{H}^{(l+1)} = {ReLU}\left( \mathbf{D}^{-\frac{1}{2}} \mathbf{\hat{A}} \mathbf{D}^{-\frac{1}{2}} \mathbf{H}^{(l)} \mathbf{W}_l \right)$
where $\mathbf{\hat{A}}=\mathbf{A} + \mathbf{I}$ denotes the adjacency matrix with the self-loop, $\mathbf{D}$ is a diagonal matrix which has its $(i,i)^{th}$ element as $\mathbf{D}_{ii} = \sum_j \mathbf{A}_{i,j}$ where $\mathbf{A}_{i,j}$ is an $(i,j)^{th}$ entry of $\mathbf{A}$, $\mathbf{H}^{(l)}$ is a feature matrix at $l^{th}$ layer, $\mathbf{W}_l$ is a parameter applied at $l^{th}$ layer, and ReLU denotes the rectified linear unit.
For each mode $m_i$, we obtain an embedded feature matrix $\mathbf{U}_{m_i}$ with the dimensionality of $\mathbf{U}_{m_i} \in \mathbb{R}^{n \times h_i}$, where $h_i$ is a number of embedding unit for mode $m_i$.
We concatenate the feature embedding matrices from multiple views into the final multi-modal latent feature matrix $\mathbf{Z}$, with the dimensionality of $\mathbf{Z} \in \mathbb{R}^{n \times h_f}$ where $h_f = \sum_i h_i$.

\vspace*{-0.3cm}
\subsection{Pruning nodes with reconstruction-based scoring} 
\vspace*{-0.2cm}
To identify nodes that are the most informative about the given graph, we first use a multi-view latent space $\mathbf{Z}$ obtained from the proposed multi-view embedding to embed each node. Then, we score the nodes based on their reconstruction errors, as we describe in the following paragraphs.
\vspace*{-0.3cm}
\paragraph{Task specific reconstruction}
We learn the latent space $\mathbf{Z}$ to detect the nodes to be pruned by reconstructing the adjacency matrix $\mathbf{\tilde{A}}$ as $\mathbf{\tilde{A}} = f(\mathbf{ZZ}^T)$, where $f$ is a sigmoid function, and the feature matrix $\mathbf{\tilde{X}}$ as $\mathbf{\tilde{X}} = {ReLU}(\mathbf{ZW}_{recon} + \mathbf{B})$ with the reconstruction parameter $\mathbf{W}_{recon} \in \mathbb{R}^{h_f \times d}$ and bias $\mathbf{B} \in \mathbb{R}^{n \times d}$, using the ReLU activation function. Then, the reconstruction loss $\mathcal{L}_a$ for the adjacency matrix is given as the mean of the negative log-likelihood for each edge, as in Eq.~\ref{La} (left). Further, the reconstruction loss for feature matrix $\mathcal{L}_x$ is defined as Eq.~\ref{La} (right), which is defined as a Frobenius norm on the difference between the original node features and the reconstructed ones.
\begin{equation} \label{La}
    \mathcal{L}_a = \frac{1}{N^2} \sum_{i=1}^n \sum_{j=1}^n -\left[\mathbf{A}_{i,j}\log\mathbf{\tilde{A}}_{i,j} + (1-\mathbf{A}_{i,j})\log(1-\mathbf{\tilde{A}}_{i,j})\right], \quad
    \mathcal{L}_x = \frac{1}{NF} || \mathbf{X} - \mathbf{\tilde{X}} ||_F^2
\end{equation}
\noindent The final reconstruction loss $\mathcal{L}_r$ is the sum of the reconstruction errors for the adjacency and the feature matrix: $\mathcal{L}_r = \mathcal{L}_a + \mathcal{L}_x$
This reconstruction loss is included in the loss for each layer, which in sum is combined with the task loss.

\vspace*{-0.2cm}
\paragraph{Scoring for node pruning}
When learning to reconstruct a given graph, along with the task loss, nodes that are less relevant for the given task will have smaller contributions when building the latent space $\mathbf{Z}$. Thus we calculate the pruning score of each node as the difference between the original and reconstructed connectivity and features for each node. In Eq.~\ref{score}, the value $s(v_i)$ represents the pruning scores of each node $v_i$.
\begin{equation} \label{score}
    s(v_i) = \lambda ||\mathbf{a}_i - \mathbf{\tilde{a}}_i ||_2^2 + (1 - \lambda) ||\mathbf{x}_i - \mathbf{\tilde{x}}_i||_2^2
\vspace*{0.1cm}
\end{equation}
where $\mathbf{a}_i(\mathbf{\tilde{a}}_i)$ and $\mathbf{x}_i(\mathbf{\tilde{x}}_i)$ indicates $i^{th}$ row of adjacency matrix and feature matrix, which corresponds to the connectivity and features for $v_i$, and $\lambda$ is a trade-off factor.
We build an indicator $\mathbf{I} \in \{0,1\}^n$ to denote which nodes to preserve based on the pruning score computed for each node. The $i^{th}$ element of $\mathbf{I}$ is computed as follows:

\vspace*{-0.35cm}

\begin{equation} \label{indicator}
\mathbf{I}_i =\begin{cases}
1, &\mbox{if }f(-s(v_i) + \mu + 2 \sigma) \ge 0.5 \\
0, &\mbox{if }f(-s(v_i) + \mu + 2 \sigma) < 0.5
\end{cases}
\end{equation}
where $\mu$ is the mean value for all pruning scores of the graph, $\sigma$ is the standard deviation for all pruning scores of the graph, and $f$ is a sigmoid function.
As shown in Eq.~\ref{indicator}, the node with its pruning score in $\mu + 2\sigma$ for its distribution is considered as an \emph{anomaly}. 
Since the pruning score is in descending order of the anomalous degree of each node, outliers can be removed by pruning nodes that have a score more than twice the standard deviation above the mean in the distribution.
The zero value in the indicator means that the corresponding node should be pruned and its feature and connectivity should not be considered at the next layers\footnote{Note that this pruning function does not affect differentiability as we round the sigmoid output with ”tf.round” function which is treated as constant when calculating gradient.}. 
This pruning policy allows the model to adaptively prune the nodes based on the input graph. If all nodes of a graph have equally low pruning scores, all nodes will be retained to be considered in the next layers. However, for any nodes that have high pruning scores, the pruning indicator vector will be set to 1 to prune them out. Such adaptive pruning is more flexible compared to strategies that drop a fixed number/ratio of nodes, which are suboptimal.

\vspace*{-0.2cm}
\subsection{Pruning and pooling} 

\vspace*{-0.1cm}
\paragraph{Graph pruning}
Using the constructed indicator, 
we obtain the pruned feature matrix, $\mathbf{X'}$ and pruned adjacency matrix, $\mathbf{A'}$ based on the indicator $\mathbf{I}$ obtained through scoring nodes for pruning:
\begin{equation} \label{pruned XA}
    \mathbf{X'} = (\mathbf{X}^{T} \cdot \mathbf{I})^{T}; \quad \mathbf{A'} = (\mathbf{A}^{T} \cdot \mathbf{I})^{T} \cdot \mathbf{I}
\end{equation}
Now, the feature matrix and adjacency matrix contain only meaningful components and ignore the anomaly features and connectivity, so that build the pruned graph clearly. 
Then we can input the pruned graph defined by the $\mathbf{X'}$ and $\mathbf{A'}$ into the pooling layer.

\vspace*{-0.2cm}
\paragraph{Combined Objective}
To train the overall model in an end-to-end manner, we also include the final task loss, which is a cross-entropy loss $\mathcal{L}_{ce}$, the reconstruction loss $\mathcal{L}_r$, then apply the $\mathcal{L}_{pool}$ from the pooling technique which MVP adopts. Thus, we finally define the combined objective as the sum of the three loss functions as shown in Eq.~\ref{total_loss}. 
\begin{equation} \label{total_loss}
    \mathcal{L} = \mathcal{L}_r  + \mathcal{L}_{ce} + \mathcal{L}_{pool}
\end{equation}\vspace*{-0.2em}
Minimizing the combined objective will make the latent space $\mathbf{Z}$ used for anomaly detection and graph pruning to be appropriately learned for the given task, such that we can select the most informative nodes in a task- and instance-adaptive manner. 

\vspace*{-0.2cm}
\subsection{Complexity Analysis}
\vspace*{-0.1cm}
The spatial complexity of the proposed method is $\mathcal{O}(Nh_f)$ where the $h_f$ denotes the number of features of latent space and $N$ is the number of nodes. Indicator consumes much less memory as it consists of a single vector for a graph with $N$ nodes, that is, $\mathcal{O}(N)$.
The computational complexity is dominated by the cost of the node embedding part. This part has the complexity of $\mathcal{O}(N^2f' + Nf'h_f)$ where f' is a maximum number of features of embedding vector for the input feature. Since the connectivity for the graph is usually sparse, we can exploit operations for sparse tensor and can reduce the complexity for the first term into $\mathcal{O}(Ef')$ where E denotes the number of edge for the graph.  Let $n$ be a number of view, then the final computational complexity is $\mathcal{O}(nf'(E + Nh_f))$.

\section{Experiment}

We evaluate our model on multiple benchmark datasets for graph classification. We also conduct various quantitative and qualitative analyses including an ablation study of the proposed method. For MVP, we use pretrained pooling weights to more clearly observe MVP's effectiveness.


\vspace*{-0.2cm}
\subsection{Supervised Graph Classification}
\definecolor{aliceblue}{rgb}{0.94, 0.97, 1.0}
\begin{table}[bt!]
\caption{\small Graph Classification Accuracy of various models and with MVP with various datasets. Each score represents the average classification accuracy $\pm$ standard deviation for 10 trials. The bold number represents the highest accuracy achieved among the methods for each dataset.}
\small
    \centering
    \resizebox{\textwidth}{!}{
    \begin{tabular}{l c c c c c c c c c}
        \thickhline
        \\[-1em]
         & \multicolumn{5}{c}{Biochemical Domain} & \multicolumn{3}{c}{Social Domain} & \multirow{2}{*}{ \makecell{ Average \\ Accuracy}}
         \\\cmidrule(lr){2-6}\cmidrule(lr){7-9}
                  & PROTEINS & DD   & FRANKENSTEIN & NCI1& HIV & COLLAB & IMDB-B & IMDB-M\\
        \hline
        \\[-1em]
        \# graphs   & 1113     & 1178 & 4337        & 4110 & 41127 & 5000   & 1000        & 1500  \\
        \# classes  & 2        & 2    & 2           &  2  & 2 & 3      & 2           & 3     \\
        Avg \# nodes& 39.06    &284.32& 16.90       &29.87& 26.0 & 74.49  & 19.77       & 13.00 \\
        \# features & 82       & 101  & 780         &  43 & 21 & 370    & 66          & 60    \\
        \\[-1em]
        \hline
        \\[-1em]
        TopKPool  & 71.16$_{\pm 4.43}$ & 69.41$_{\pm 6.38}$ & 62.47$_{\pm 3.98}$
                  & 61.09$_{\pm 8.03}$
                  & 72.18$_{\pm 4.32}$
                  & 78.64$_{\pm 1.51}$ & 68.22$_{\pm 10.28}$ & 48.48$_{\pm 8.05}$ & 66.46 
                    \\
        SAGPool    & 70.80$_{\pm 5.22}$ & 68.39$_{\pm 7.86}$ & 62.97$_{\pm 3.31}$
                  & 65.94$_{\pm 8.31}$
                  & 68.49$_{\pm 5.30}$
                  & 82.20$_{\pm 1.36}$ & 69.56$_{\pm 8.71}$ & 49.40$_{\pm 5.59}$ & 67.22
                    \\
        MLP      & 70.00$_{\pm 4.23}$ & 71.69$_{\pm 2.71}$ & 58.16$_{\pm 1.44}$
                  & 65.21$_{\pm 3.27}$
                  & 67.08$_{\pm 3.11}$
                  & 79.90$_{\pm 1.27}$ & 71.20$_{\pm 2.56}$ & 50.13$_{\pm 1.14}$ & 66.55
                    \\
        \rowcolor{aliceblue}
         MVP + MLP & 74.20$_{\pm 3.54}$ & 74.24$_{\pm 3.13}$ & 59.03$_{\pm 2.59}$
                  & 63.38$_{\pm 1.78}$
                  & 67.86$_{\pm 2.37}$
                  & 81.58$_{\pm 2.09}$ & 73.60$_{\pm 3.01}$ & 53.00$_{\pm 2.77}$ & 67.95  \\
        GCS & 72.59$_{\pm 3.70}$ & 75.08$_{\pm 3.54}$ & 65.07$_{\pm 2.64}$
                  & 73.02$_{\pm 2.80}$
                  & 72.84$_{\pm 2.16}$
                  & 81.18$_{\pm 2.20}$ & 76.20$_{\pm 4.12}$ & 50.13$_{\pm 3.50}$ & 70.76  \\
        \rowcolor{aliceblue}
        MVP + GCS & 75.80$_{\pm 4.27}$ & 77.03$_{\pm 3.27}$ & 68.13$_{\pm 2.68}$
                  & 73.08$_{\pm 2.71}$
                  & \textbf{78.38}$_{\pm 4.11}$
                  & 82.22$_{\pm 2.44}$ & 76.30$_{\pm 4.45}$ & 52.93$_{\pm 4.21}$ & 72.77  \\
        DiffPool   & 73.12$_{\pm 4.51}$ & 75.34$_{\pm 3.48}$ & 58.25$_{\pm 6.10}$
                  & 74.98$_{\pm 1.56}$
                  & 71.62$_{\pm 2.63}$
                  & 81.57$_{\pm 3.98}$ & 71.98$_{\pm 5.69}$ & 51.87$_{\pm 3.14}$ & 69.84
                    \\
        \rowcolor{aliceblue}
         MVP + Diff& 79.91$_{\pm 4.38}$ & 76.16$_{\pm 3.04}$ & 68.13$_{\pm 3.23}$
                  & 75.43$_{\pm 2.44}$
                  & 76.34$_{\pm 4.17}$
                  & 83.86$_{\pm 0.87}$ & 73.30$_{\pm 2.49}$ & 52.40$_{\pm 2.09}$ & 73.19  \\
        MinCutPool & 76.87$_{\pm 2.92}$ & 78.56$_{\pm 3.10}$ & 64.56$_{\pm 3.03}$
                  & 76.20$_{\pm 1.79}$
                  & 72.19$_{\pm 4.65}$
                  & 81.38$_{\pm 1.86}$ & 72.30$_{\pm 2.45}$ & 50.13$_{\pm 3.54}$ & 71.52
                    \\
        \rowcolor{aliceblue}
        MVP + MinCut
            & \textbf{81.70}$_{\pm 3.10}$ & \textbf{82.88}$_{\pm 2.42}$
            & \textbf{68.73}$_{\pm 3.62}$ & \textbf{78.08}$_{\pm 2.68}$
            & 76.39$_{\pm 2.65}$
            & \textbf{83.92}$_{\pm 1.33}$
            & \textbf{76.70}$_{\pm 3.52}$ & \textbf{54.60}$_{\pm 5.18}$ & \textbf{75.33}     
            \\
        GMT & 77.59$_{\pm 3.00}$ & 78.72$_{\pm 4.40}$ & 61.36$_{\pm 1.95}$
                  & 66.40$_{\pm 4.68}$
                  & 61.43$_{\pm 5.39}$
                  & 80.05$_{\pm 1.71}$ & 73.80$_{\pm 4.02}$ & 50.40$_{\pm 3.48}$ & 68.72
        \\
        \rowcolor{aliceblue}
        MVP + GMT & 79.46$_{\pm 2.22}$ & 79.31$_{\pm 2.83}$ & 62.03$_{\pm 3.52}$ & 73.23$_{\pm 2.71}$
                  & 66.89$_{\pm 5.43}$
                  & 81.27$_{\pm 1.85}$
                  & 74.40$_{\pm 4.76}$  & 50.80$_{\pm 3.44}$ & 69.64 
        \\
        \thickhline
    \end{tabular}
    }
    \label{tab:acc}
    \vspace*{-0.4cm}
\end{table}

\begin{figure}
    \centering
    \subfigure[\label{fig:avgnodeperperf} Pruning efficacy]{\includegraphics[height=.21\textwidth]{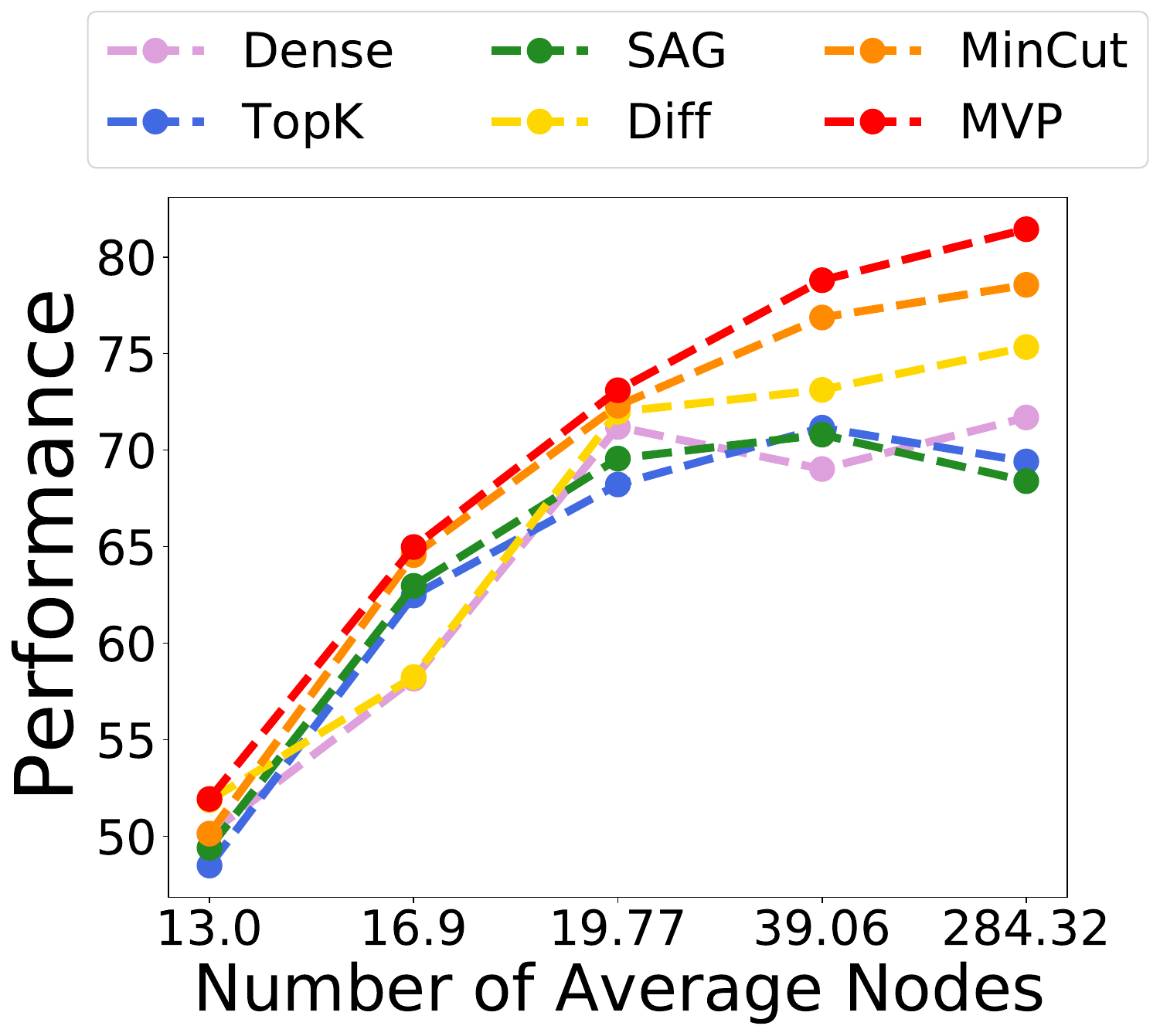}}
    \subfigure[\label{fig:prune_ratio}Pruning degree]{\includegraphics[height=.21\textwidth]{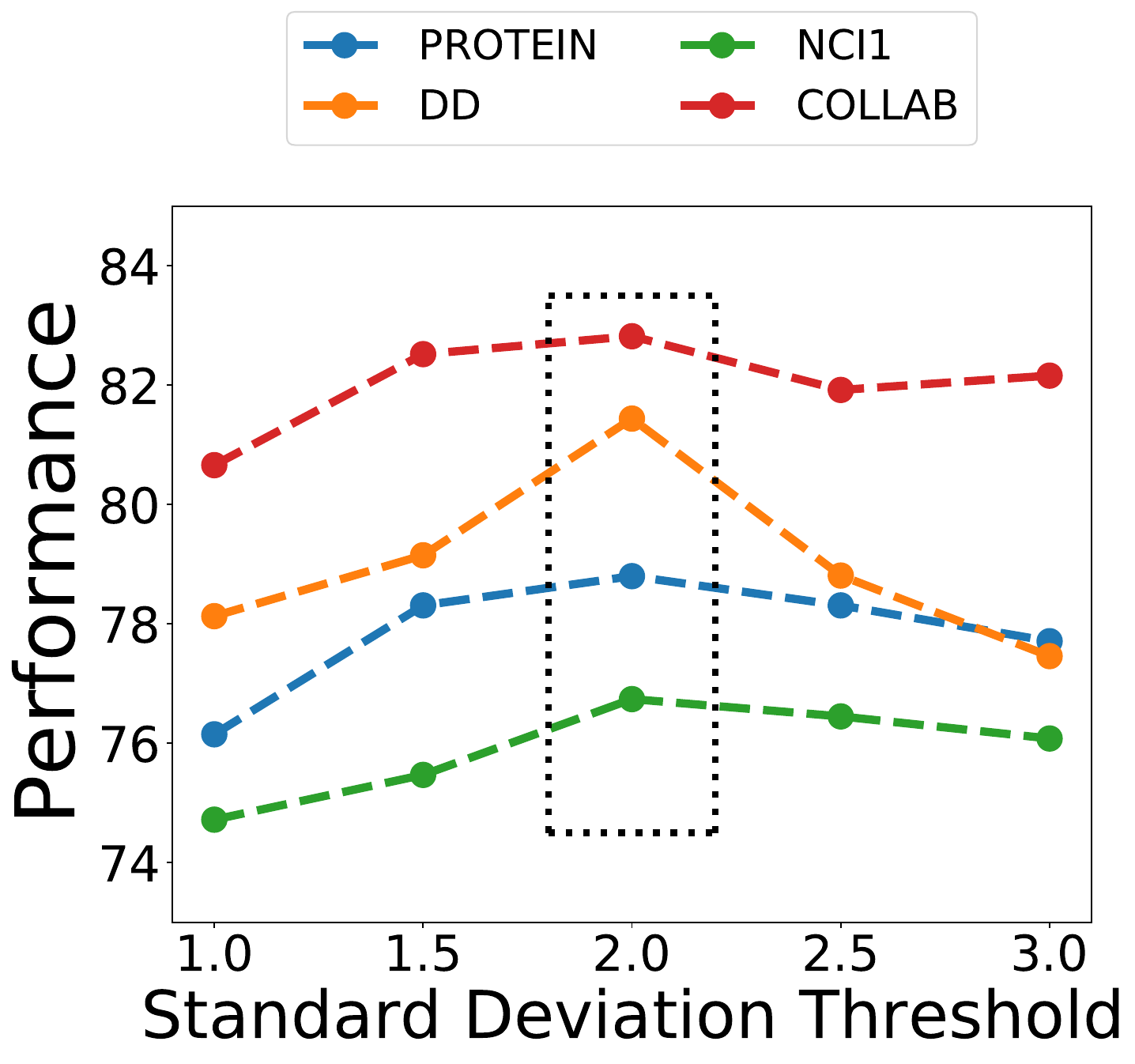}}
    \subfigure[\label{fig:param_test}Parameter test]{\includegraphics[height=.21\textwidth]{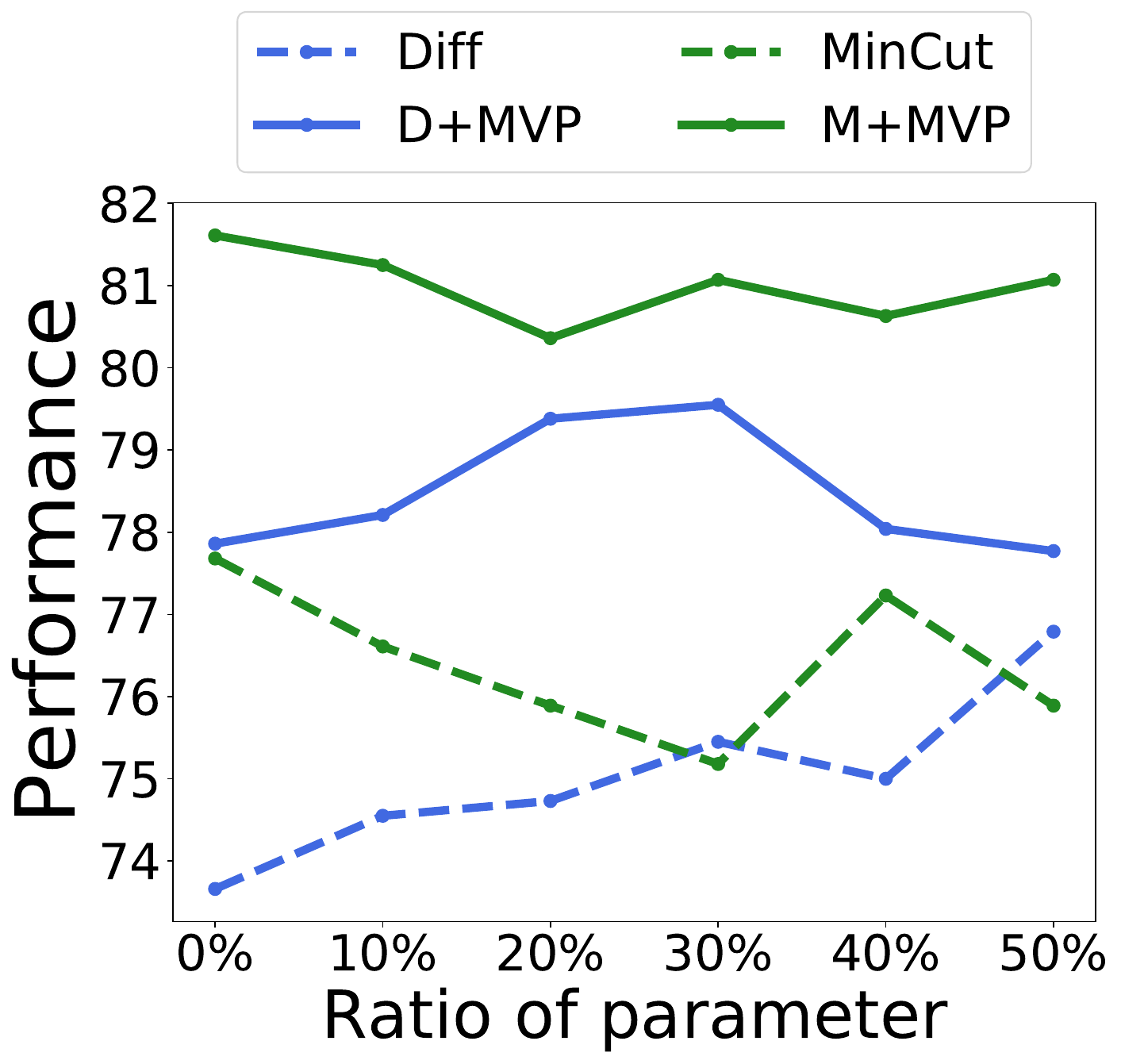}}
    \subfigure[\label{fig:mem}Memory Usage]{\includegraphics[height=.21\textwidth]{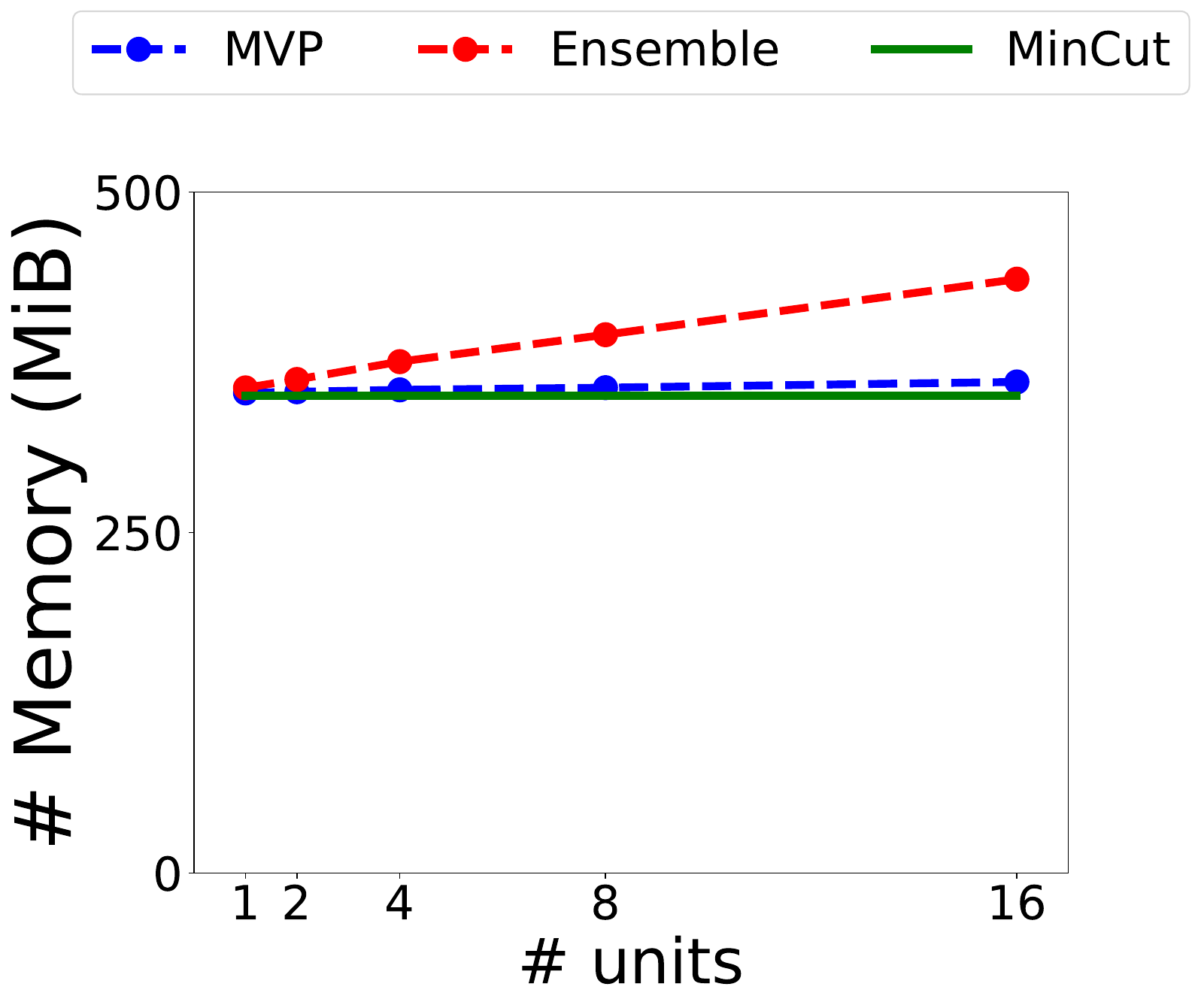}}

    \vspace*{-0.1cm}
    \caption{\small (a) The accuracy of the each models per number of average nodes in each dataset. (b) The performance of each dataset for each pruning degree. (c) The performance for each parameter ratio of PROTEIN dataset. (d) Memory usage of models for each number of view or model.}
    \label{fig:my_label}
    \vspace*{-0.5cm}
\end{figure}

\vspace*{-0.2cm}
We first validate the graph classification performance of our model on various benchmark datasets against existing graph pooling methods. We randomly split each dataset into the training set, validation set, and test set with 81\%, 9\%, 10\% ratio respectively. At this time, the degree of overlap was defined for each view so that if the number of features is insufficient, it can be compensated. Overlapping is to share features equally between adjacent views at a certain ratio. We run each model 10 times and report the average performances as well as the standard errors. The description of the datasets and baselines used are given below. 
\vspace*{-0.2cm}
\paragraph{Datasets}
%
Among the TU datasets\footnote{https://ls11-www.cs.tu-dortmund.de/staff/morris/graphkerneldatasets}, we select 7 datasets including 4 datasets from the biochemical domain (PROTEINS~\cite{borgwardt2005protein, dobson2003distinguishing}, DD~\cite{dobson2003distinguishing,shervashidze2011weisfeiler}, FRANKENSTEIN~\cite{orsini2015graph}, NCI1~\cite{wale2008comparison}), and 3 datasets from the social network domain (COLLAB~\cite{yanardag2015deep}, IMDB-BINARY~\cite{rossi2015network}, IMDB-MULTI~\cite{rossi2015network}). We also included HIV~\cite{hu2021open} dataset from OGB datasets\footnote{https://ogb.stanford.edu/docs/graphprop/} to show our method's performance on large scale graph datasets. We provide basic statistics of each dataset in the upper rows of Table \ref{tab:acc}, and provide detailed descriptions of each dataset in Section \ref{supple:dataset} of the supplementary file.

\vspace*{-0.2cm}
\paragraph{Baselines}
We compare the performance of our model against various pooling baselines. Specifically, we compare against three graph pooling methods based on node coarsening, namely GMT~\cite{baek2021accurate}, MinCutPool~\cite{bianchi2020spectral} and DiffPool~\cite{ying2018hierarchical}, and two based on node pruning, which are  SAGPool~\cite{lee2019self} and TopKPool~\cite{gao2019graph}. We also compare against a multi-layer perceptron (MLP). For a fair comparison, we conduct experiments with the same number of layers (2) for all models, including ours.

\vspace*{-0.2cm}
\paragraph{Classification Performance}
Table~\ref{tab:acc} shows that our model, MVP, largely outperforms existing pooling baselines on all datasets, including the state-of-the-art method.
Figure~\ref{fig:avgnodeperperf} shows that datasets that contain graphs with a large number of nodes on average, such as PROTEINS, DD, and COLLAB, our method achieves larger gains.
However, the performance of pruning-based pooling models, such as SAGPool and TopKPool, significantly degraded the dataset containing larger graphs. For MVP, we set the pruning threshold as twice the standard deviation, as we empirically achieve the best performance on graphs with varying sizes, as shown in Figure~\ref{fig:prune_ratio}.

\vspace*{-0.2cm}
\paragraph{Parameter and memory overhead}
We verify that MVP does not simply obtain improved performance by using more parameters. We compared the accuracy of MVP and SOTA baselines with a similar number of parameters by increasing their number of parameters by 10\%-50\% (See Figure~\ref{fig:param_test}). Although this yields marginal improvements in DiffPool, the performance gap between DiffPool + MVP and the base DiffPool is significantly large, regardless of the number of parameters.
We also show that our method results in marginal memory overhead over the base model in Figure \ref{fig:mem}, while a simple ensemble model requires significantly larger memory. We provide a more detailed figure on this experiment in Section \ref{supple:param} of the supplementary file.
\begin{table}[bt!]
\caption{\small Comparison of Multiview and ensemble model. Multi-view is significantly outperforms ensemble model in each number of views(models).}
\resizebox{\textwidth}{!}{
\begin{tabular}{ c c c c c c c c }
\thickhline
\multirow{2}{*}{Scoring policy} &  \multicolumn{3}{c}{\begin{tabular}[c]{@{}l@{}}Ensemble model\end{tabular}} &
\multirow{2}{*}{\begin{tabular}[c]{@{}l@{}}Single-view\\ model\end{tabular}} & \multicolumn{3}{c}{\begin{tabular}[c]{@{}l@{}}Multi-view model\end{tabular}} \\ 
\cmidrule(lr){2-4} \cmidrule(lr){6-8}
                    & 2 model & 4 model & 8 model &        & 2 view & 4 view & 8 view \\\hline
MVP scoring         &  79.64$_{\pm 5.34}$  &  79.76$_{\pm 3.50}$  &  79.69$_{\pm 4.80}$  & 79.46$_{\pm 3.50}$  & 80.09$_{\pm 3.68}$  & 80.71$_{\pm 2.88}$  & 81.70$_{\pm 3.10}$ \\
Attention scoring   &  -      &  -      &  -      & 69.37$_{\pm 5.10}$  & 68.12$_{\pm 4.28}$  & 69.55$_{\pm 4.58}$  & 71.34$_{\pm 3.76}$ \\ \thickhline
\end{tabular}
}
\label{tab:ensemble}
\vspace*{-0.5cm}
\end{table}

\vspace*{-0.2cm}
\subsection{Analysis of Multi-view Embedding}
\vspace*{-0.2cm}
We empirically demonstrated the effectiveness of our multi-view embedding method by comparing it to SAGPool which uses node scoring policy as attention-based scoring. As shown in Table~\ref{tab:ensemble}, considering multiple view improves performance over consideration of single view for both methods, although MVP largely outperforms SAGPool as it also uses reconstruction-based node scoring. Table~\ref{tab:ensemble} also shows that our multi-view embedding method is not a simple ensemble method, since using ensemble models yields marginal performance gains, that are significantly smaller than gains using multi-view embeddings.

\vspace*{-0.2cm}
We further verify the effectiveness of multi-view embedding by examining the importance score of each feature in Figure~\ref{fig:importance_score}, which are computed using the Local Interpretable Model-agnostic Description model~\cite{ribeiro2016should} which estimates the degree to which small changes in variable affect the predicted value. In Figure~\ref{fig:importance_score}, the black and grey bar plots represent the feature importance for the single-view model, and the colored bar plot represents the feature importance for the multi-view model, where we use different colors to denote each view. As shown in this figure, when a model is trained without considering multiple views, it pays attention only to the dominant features, which are mostly topological features.

\vspace*{-0.2cm}
However, with our multi-view embedding, the model places a larger importance score on non-topological features as well. In other words, feature importance scores are distributed in a balanced way. As the multi-modality is considered for each multi-view, the performance is improved as shown in Table~\ref{tab:ensemble}.  In specific, such multi-view embedding can be done through random view splitting as well as utilizing intrinsic modalities. Combining a random subset of features may break correlation across features such that the model can learn about each view in a less biased manner.

\vspace*{-0.3cm}
\subsection{Analysis of Reconstruction-based Pruning Scoring}
\vspace*{-0.3cm}
Pruning nodes based on their reconstruction errors is one of the important technical contributions of our method. To show that our reconstruction-based scoring is critical for detecting and pruning anomalous nodes, we provide the comparison of it against the attention-based scoring, in this section. We further show that MVP prunes the nodes which are not semantically important for classifying graphs.

\begin{figure}
\centering
    \begin{minipage}[b]{.28\linewidth}
        \centering
        \includegraphics[width=\textwidth,height=8cm]{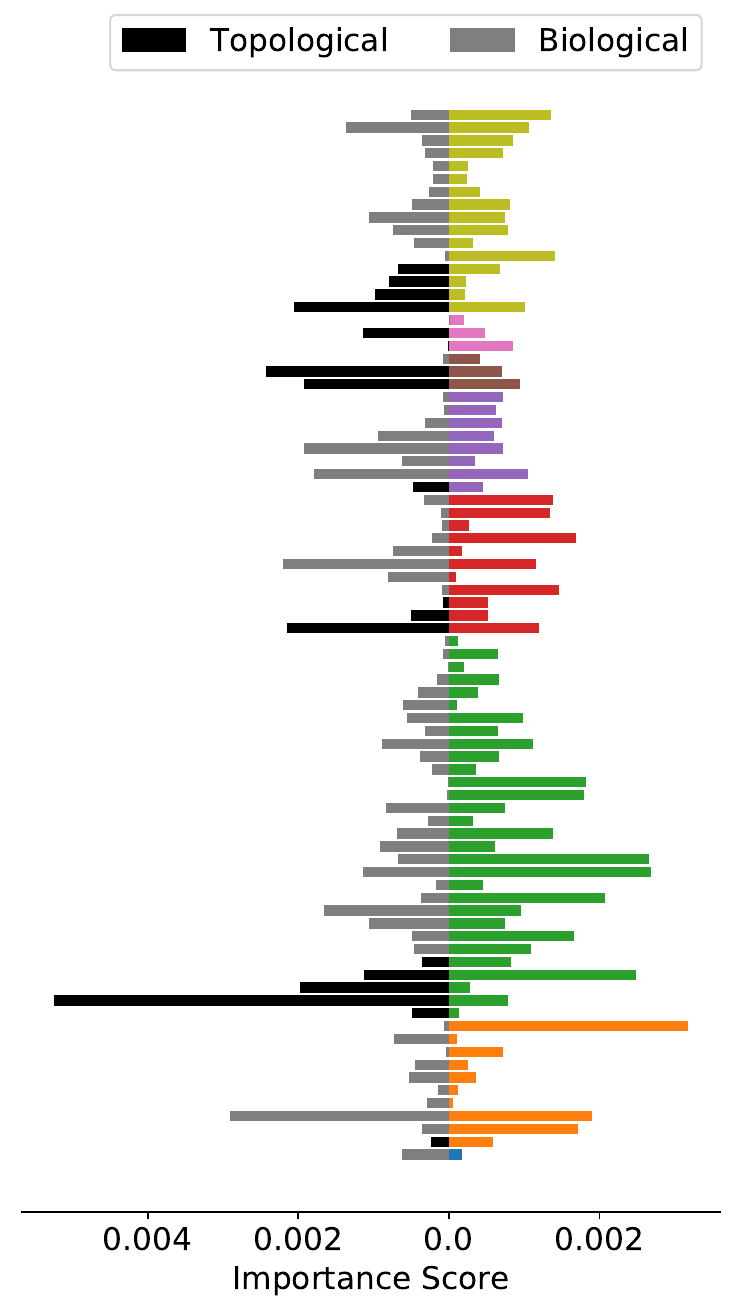}
        \caption{\small Importance score of each feature.}
        \label{fig:importance_score}
    \end{minipage}
    \hspace{0.5em}
    \begin{minipage}[b]{.68\linewidth}
        \begin{minipage}[b]{.43\linewidth} 
        \centering
        \includegraphics[height=3.3cm,width=\textwidth]{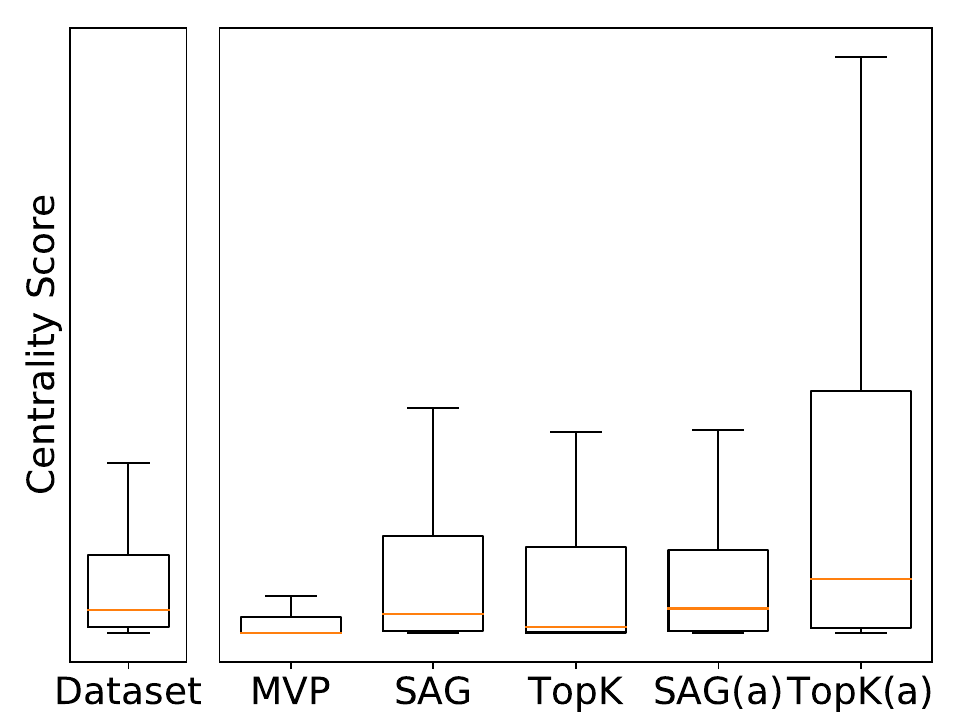}
        \parbox{\textwidth}{\caption{\label{fig:centrality} \small Boxplot of Betweenness-centrality scores distribution.}}
        \vspace*{-0.3cm}
        \end{minipage}
        \hspace{1em}
        \begin{minipage}[b]{.55\linewidth}
        \centering  
            \includegraphics[height=3.1cm, trim=0cm 1cm 8.3cm 0.5cm, clip]{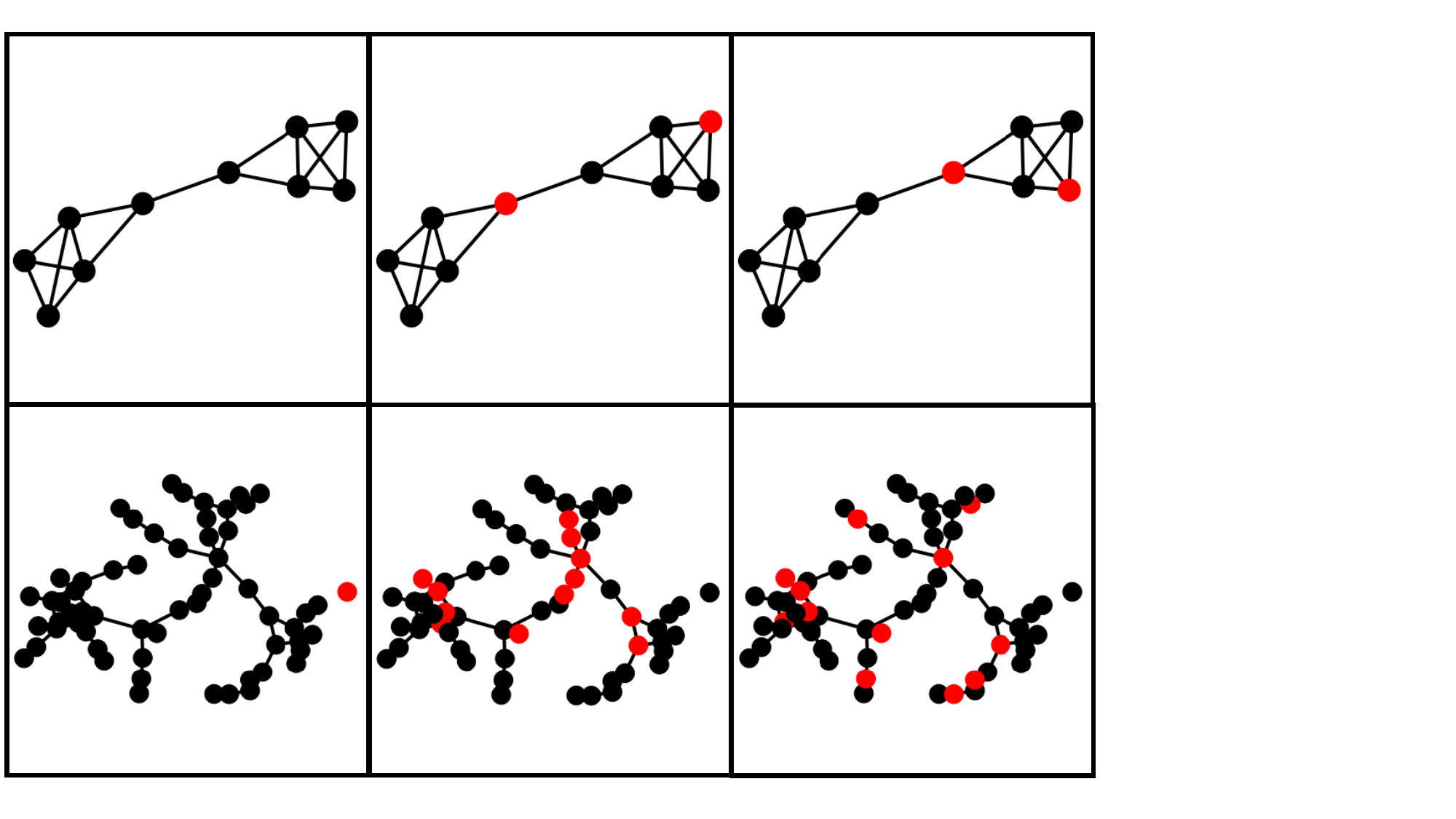}
        \vspace{-0.1cm}
        \caption{\label{fig:pruned}\small Adaptive Pruning Examples in various models. Left: MVP, Middle: SAGPool, Right: TopKPool}
        \end{minipage}
        
        \begin{minipage}[b]{.43\linewidth}
            
            \includegraphics[width=\textwidth,height=3.5cm]{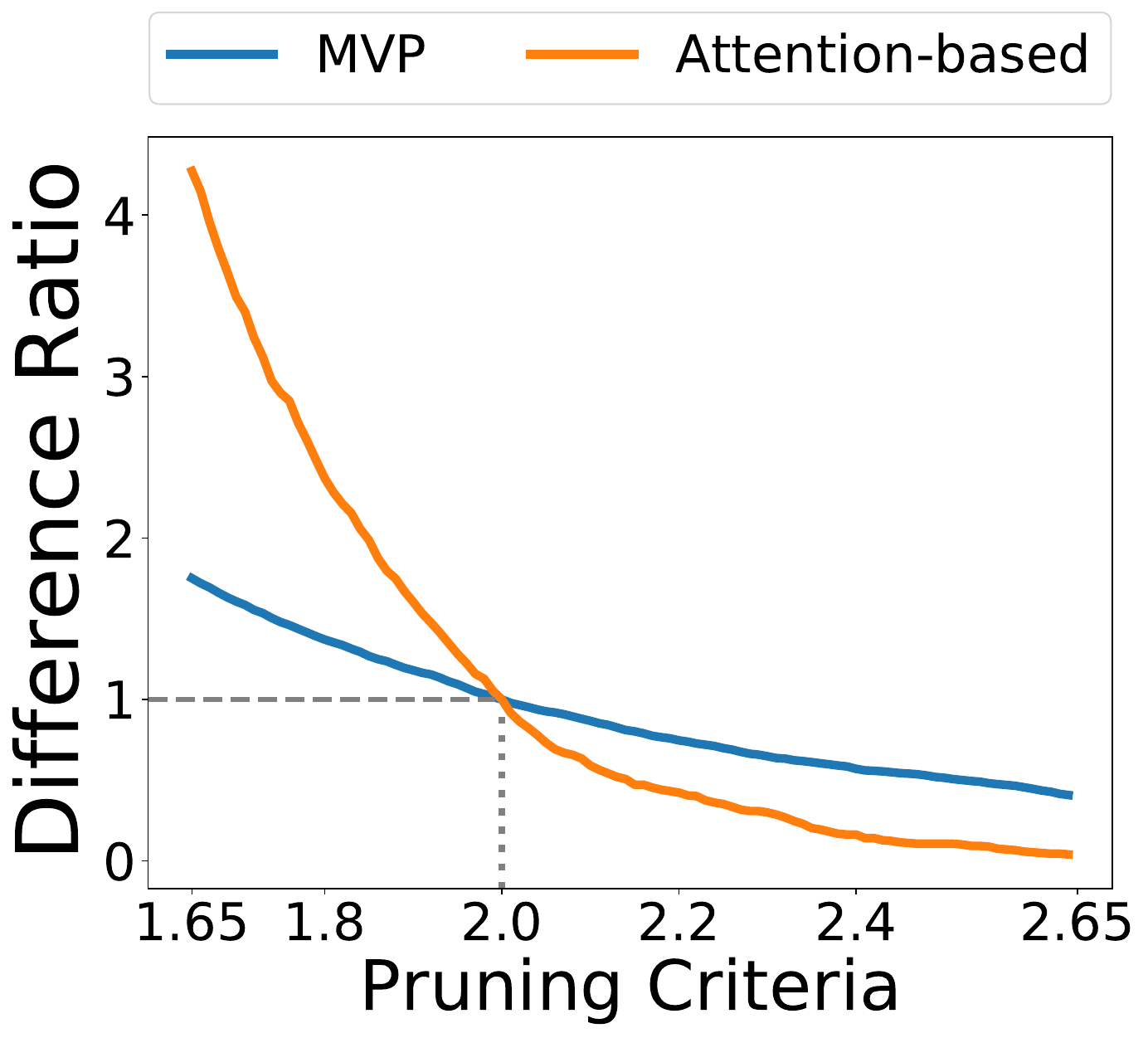}
            \vspace{-0.5cm}
            \caption{\small The number of pruning nodes according to the performance of each scoring policy.}\label{fig:pruning_ratio}
        \end{minipage}
        \hspace{1em}
        \begin{minipage}[b]{.55\linewidth}
            \centering
            \vspace{-0.2cm}
            
            \resizebox{.97\textwidth}{!}{
            \begin{tabular}{l c}
            \thickhline
            Model & Performance \\ \hline \hline
            MVP & \textbf{81.70$_{\pm 3.10}$} \\ \hline
            {\small Degree-based (10\%)} & 71.70$_{\pm 3.83}$ \\
            {\small Degree-based (20\%)} & 70.63$_{\pm 4.13}$ \\
            {\small Degree-based (< 4)} & 71.18$_{\pm 3.32}$ \\
            {\small Degree-based (< 3)} & 70.63$_{\pm 4.06}$ \\\hline
            TopKPool & 71.16$_{\pm 4.43}$ \\
            SAGPool & 70.80$_{\pm 5.22}$ \\
            \thickhline
            \end{tabular}}
            \captionof{table}{\small Pruning Performance of degree-based models and existing pruning-based pooling models with Protein data.}
            \label{tab:degree_pruning}
        \end{minipage}
    \end{minipage}
    \vspace*{-0.7cm}
\end{figure}

\vspace*{-0.2cm}
\paragraph{Scoring for pruning and centrality}
The objective of anomaly detection is pruning abnormal nodes in the graph. Thus, the model should prune the node which is less informative of the given graph. 
We assess the unnecessity of pruned nodes by calculating the betweenness centrality. Specifically, the betweenness centrality is defined as $C_B(v) = \sum_{s,v,t \in V}{ \frac{\sigma_{st}(v)}{\sigma_{st}} }$ for $\sigma_{s,t}(v)$ as the shortest path of pair of nodes $(s,t)$ which pass through $v$ where $\sigma_{s,t}$ as the all shortest path of $(s,t)$. 
The high betweenness centrality score for a node suggests that it is essential for connecting one node to another. On the contrary, a node with a low betweenness centrality score may have little contribution to propagating information and thus, may not be informative. 
Figure~\ref{fig:centrality} shows the statistics of the betweenness centrality scores of all nodes in the PROTEINS dataset and pruned nodes of each pooling method which is based on node pruning. 
We further use SAG(a) and TopK(a), modification of the SAGPool and TopKPool that use adaptively pruning as our method does, rather than pruning a fixed number of nodes or fixed ratio of nodes.
We calculate the harmonic mean value of the betweenness centrality score of each graph to consider that the number of nodes in each graph is different. As shown in the box plot, MVP tends to prune nodes with lower values compared to other models and has a near-zero median value (orange line).
This verifies that MVP prunes nodes with low betweenness centrality scores, which are less representative.

\vspace*{-0.2cm}
The qualitative examples in Figure~\ref{fig:pruned} show 
that our reconstruction-based pruning score considers both the topological features and the betweenness centrality of the graph when performing pruning.
The graphs with the pruned nodes for SAGPool and TopKPool (Middle and the Right column of Figure~\ref{fig:pruned} show that they also prune nodes with high betweenness centrality,
which may be important for propagating information across nodes, even with the adaptive policy.

\vspace*{-0.2cm}
\paragraph{Comparison against the attention scoring}
To prune the node, the MVP scoring is a more sophisticated way to score the nodes than using attention-based scoring. As shown in Figure \ref{fig:pruning_ratio}, with MVP pruning, the number of the pruned nodes does not change significantly as the pruning threshold changes. However, attention-based pruning incurs a large difference in the number of pruned nodes even with a slight change in the threshold. This is because attention-based scoring does not precisely score the nodes, leading most of the scores concentrated near the pruning threshold. Moreover, since attention-based scoring is largely affected by the node aggregation of GNN, it tends to preserve nodes with high degrees, that contain information about a larger number of nodes. As a result, nodes with relatively low degrees or side branches receive low attention scores, and are more likely to be pruned. Thus, it basically works similarly to simple degree-based pruning, as shown in Table \ref{tab:degree_pruning}, where the degree-based pruning achieves almost the same performance as the attention-based node scoring methods. Thus, MVP scoring is more suitable for pruning, as it is more robust to small changes in the threshold, and is not dominated by the node's degrees.


\vspace*{-0.2cm}
\subsection{Ablation Study and Qualitative Analysis}

\vspace*{-0.2cm}
\paragraph{Loss Ablation study}
We further show the efficacy of each loss in Figure~\ref{total_loss}, through an ablation study on the PROTEIN and DD datasets. The results of this ablation study in Table \ref{tab:loss_abl} show that the optimal loss combination includes all losses, including the task-specific loss $L_{CE}$, pooling specific loss $L_{pool}$, and reconstruction loss $L_{r}$, although the reconstruction loss is the most effective as it yields the largest performance improvement when used alone.  



\begin{figure}[bt!]
    \centering 
    \subfigure[3D figure]{\includegraphics[width=.18\columnwidth,height=.13\columnwidth]{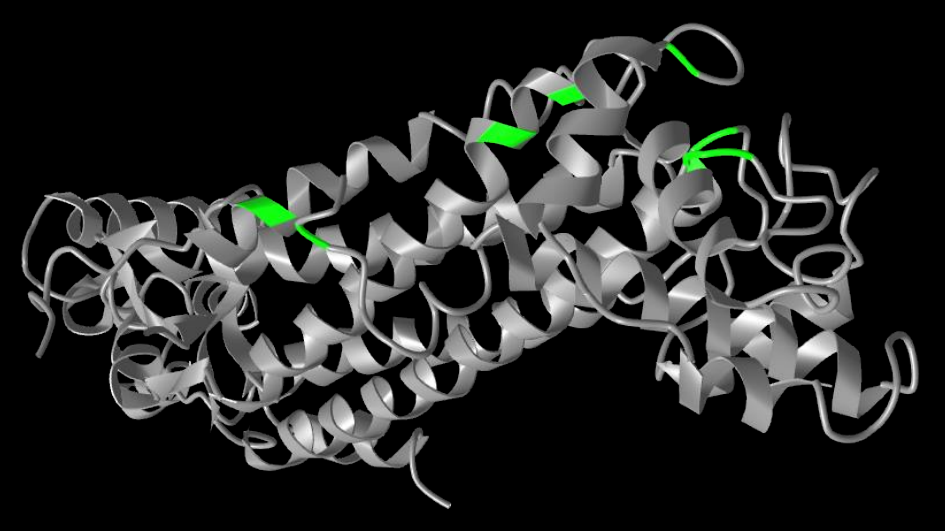}}
    \subfigure[1 view]{\centering \includegraphics[trim=0cm 2cm 6cm 2cm, clip,width=.18\linewidth]{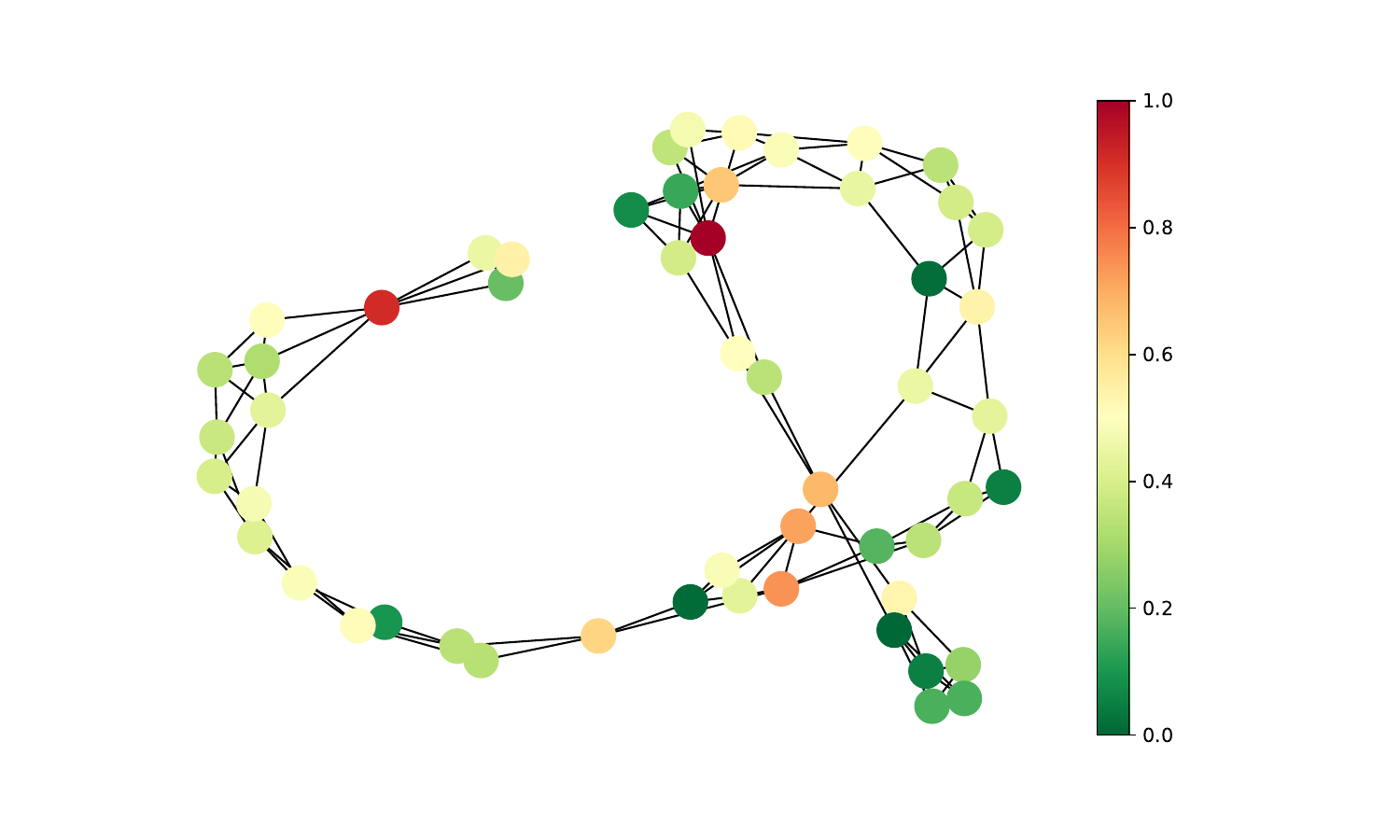}}
    \subfigure[2 view]{\centering \includegraphics[trim=0cm 2cm 6cm 2cm, clip,width=.18\linewidth]{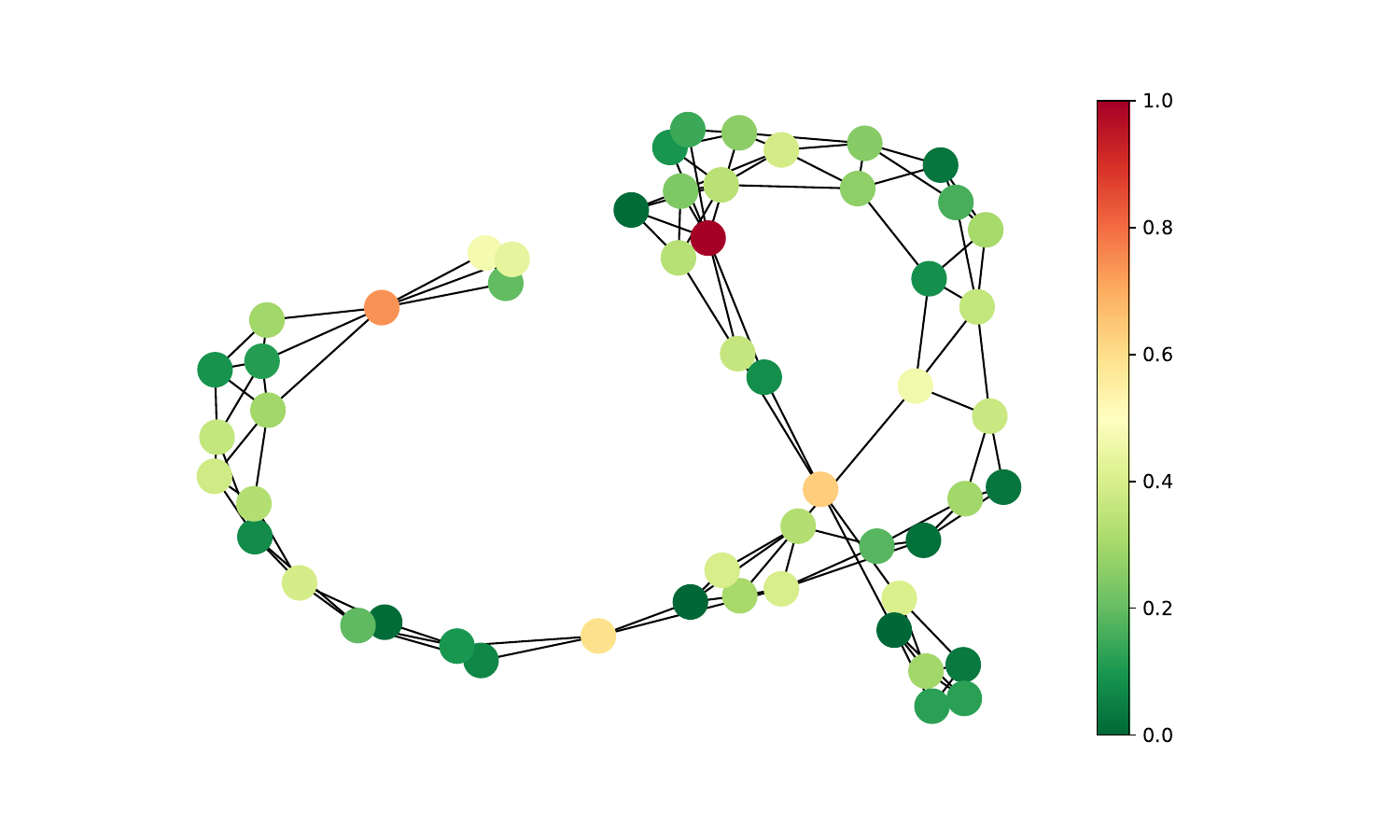}}
    \subfigure[4 view]{\centering \includegraphics[trim=0cm 2cm 6cm 2cm, clip,width=.18\linewidth]{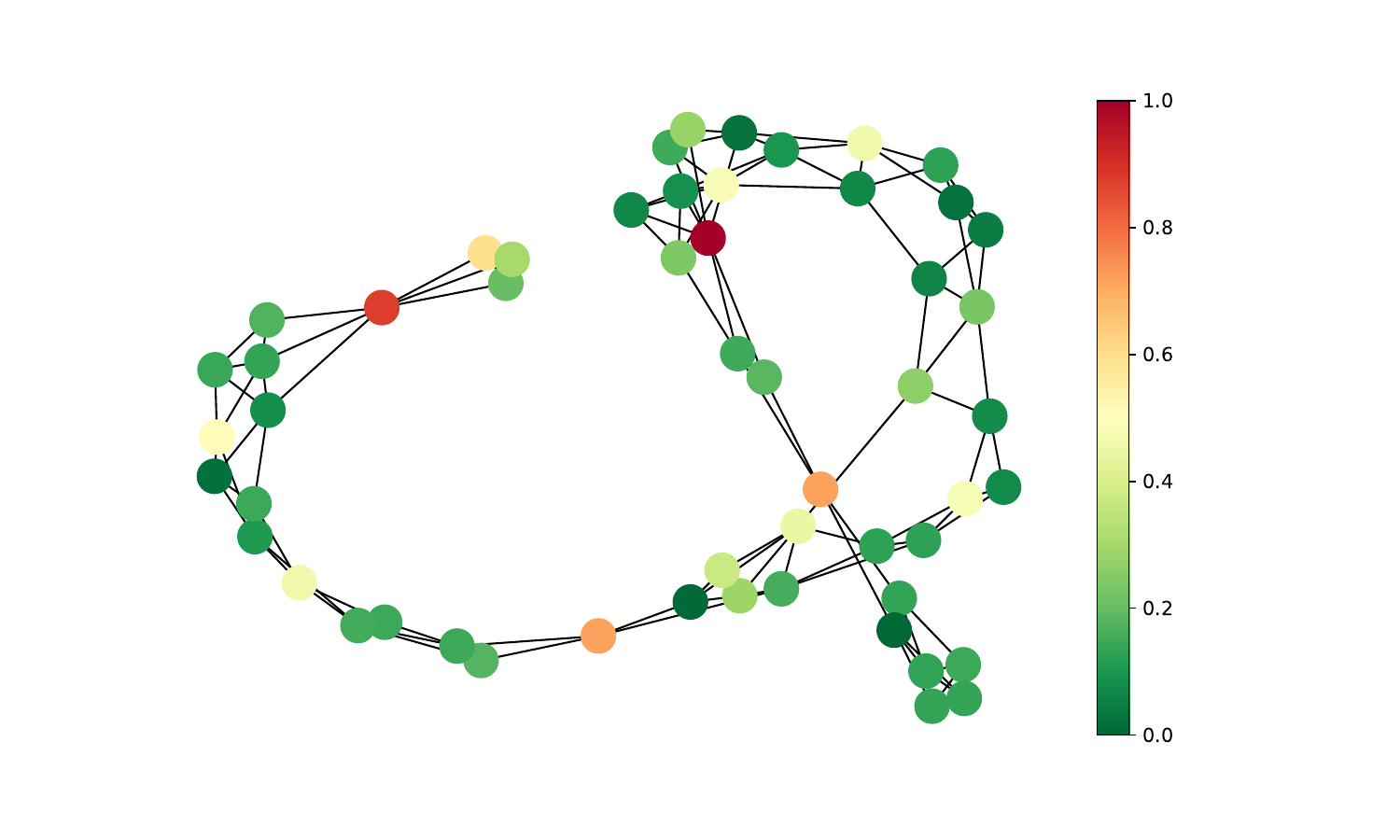}}
    \subfigure[8 view]{\centering \includegraphics[trim=0cm 2cm 5cm 2cm, clip,width=.2\linewidth]{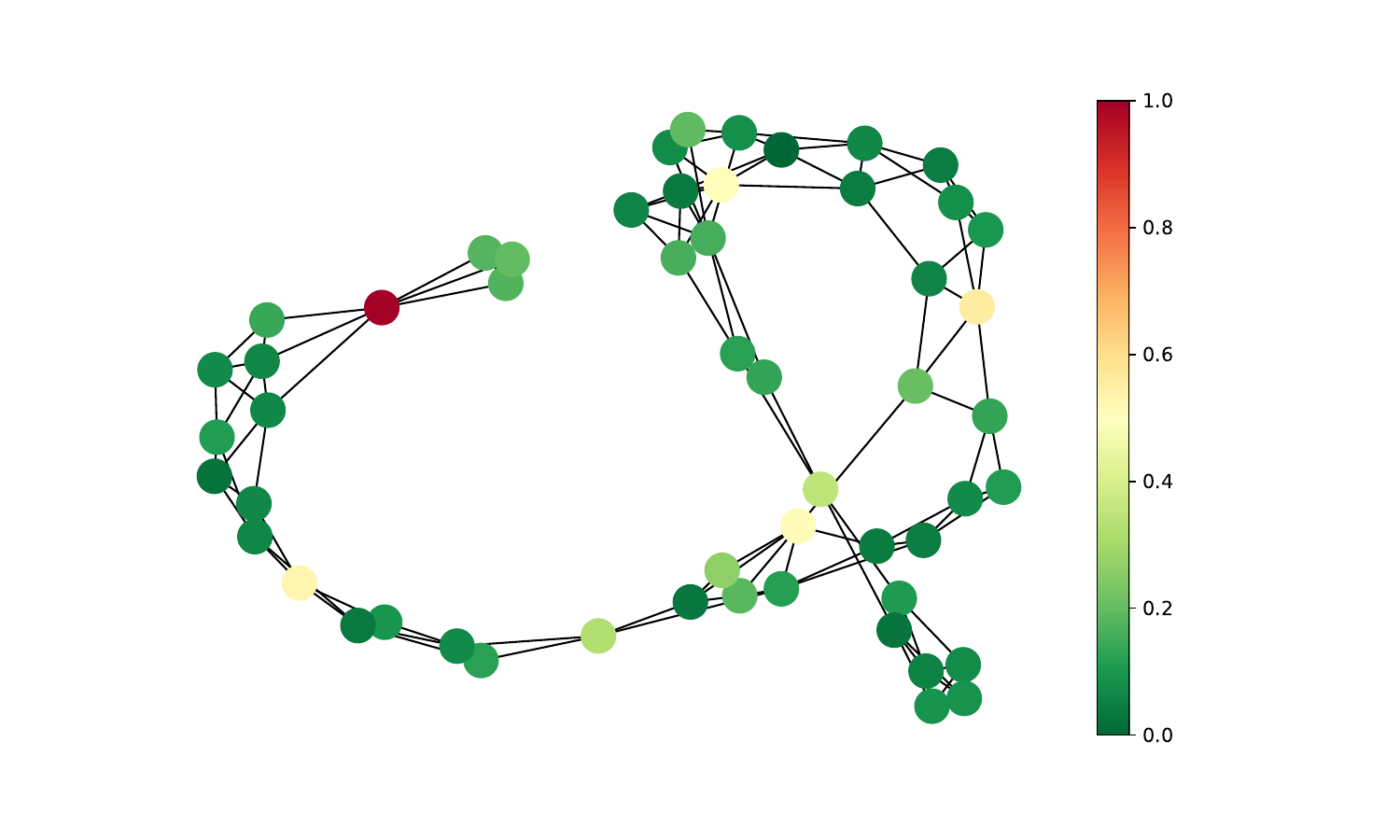}}
    
    \subfigure[critical sites]{\centering \includegraphics[trim=0cm 2cm 6cm 2cm, clip,width=.18\linewidth]{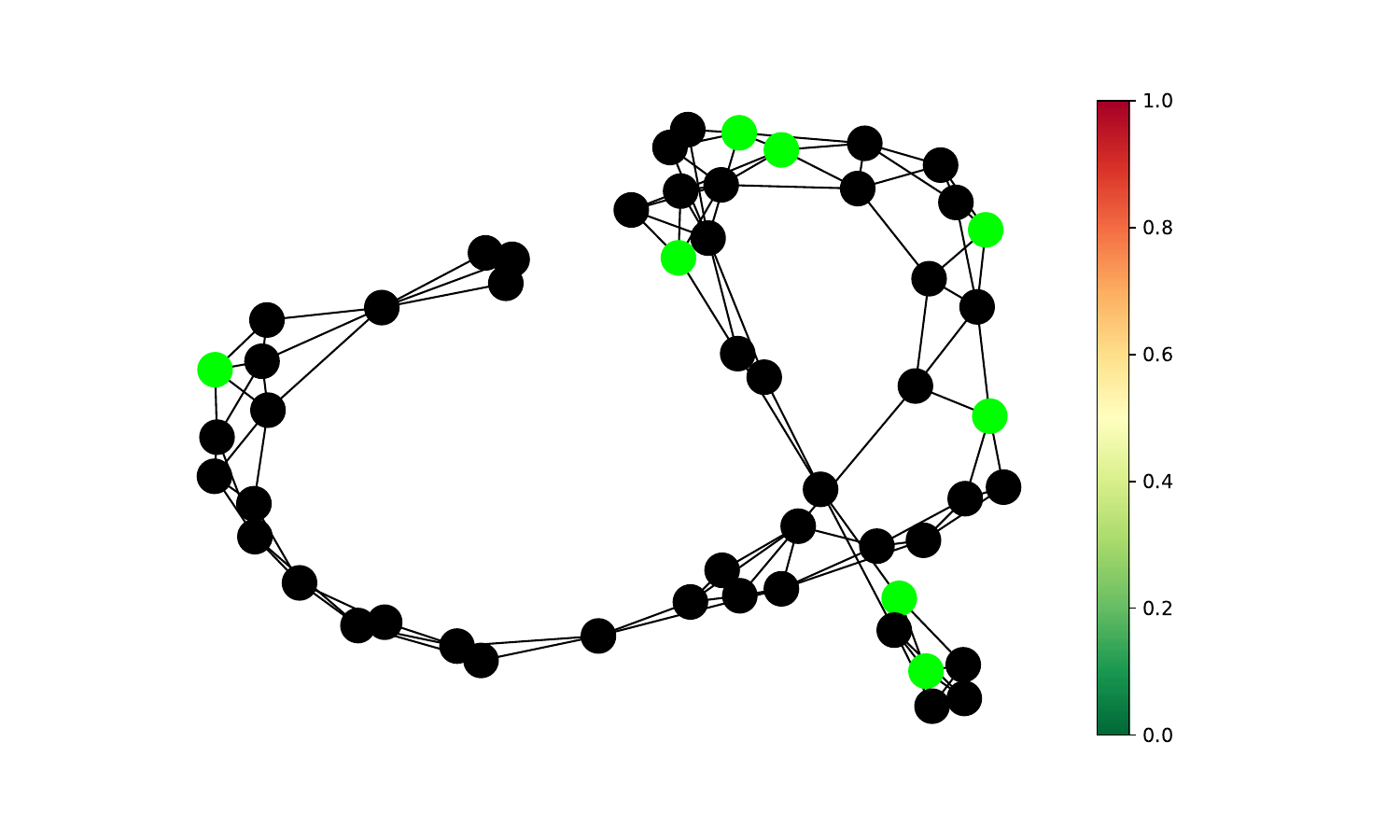}}
    \subfigure[SAGpool]{\centering \includegraphics[trim=0cm 2cm 6cm 2cm, clip,width=.18\linewidth]{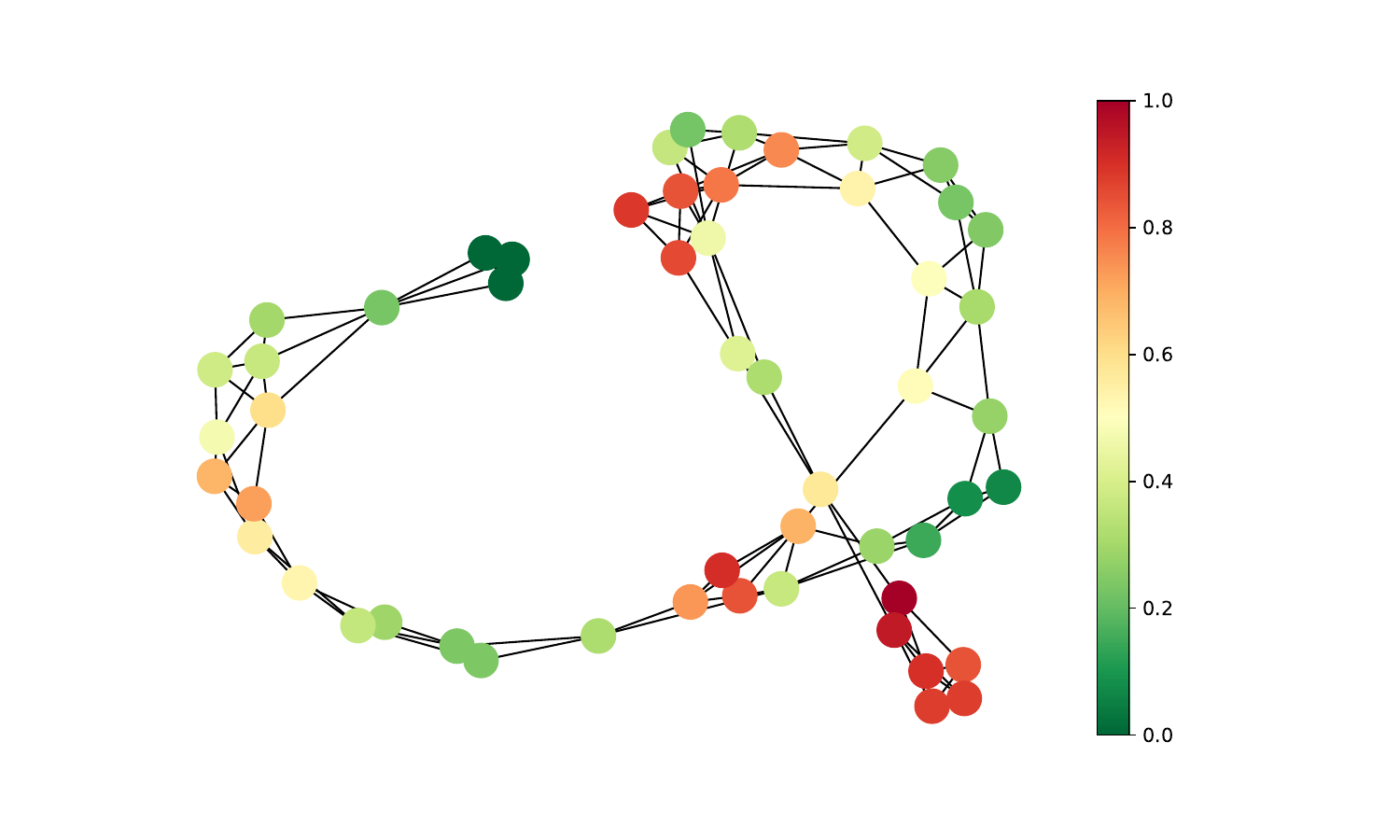}}
    \subfigure[2 models]{\centering \includegraphics[trim=0cm 2cm 6cm 2cm, clip,width=.18\linewidth]{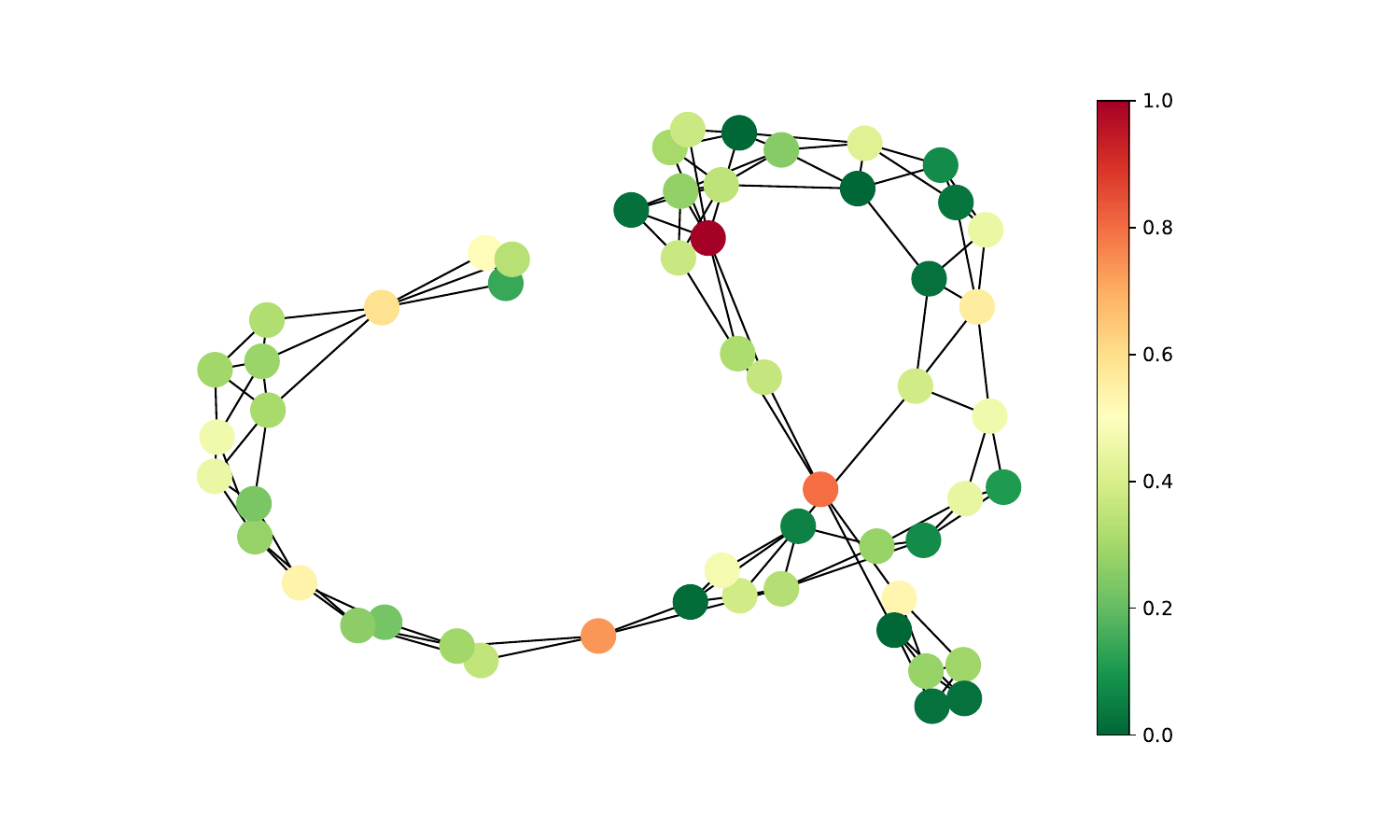}}
    \subfigure[4 models]{\centering \includegraphics[trim=0cm 2cm 6cm 2cm, clip,width=.18\linewidth]{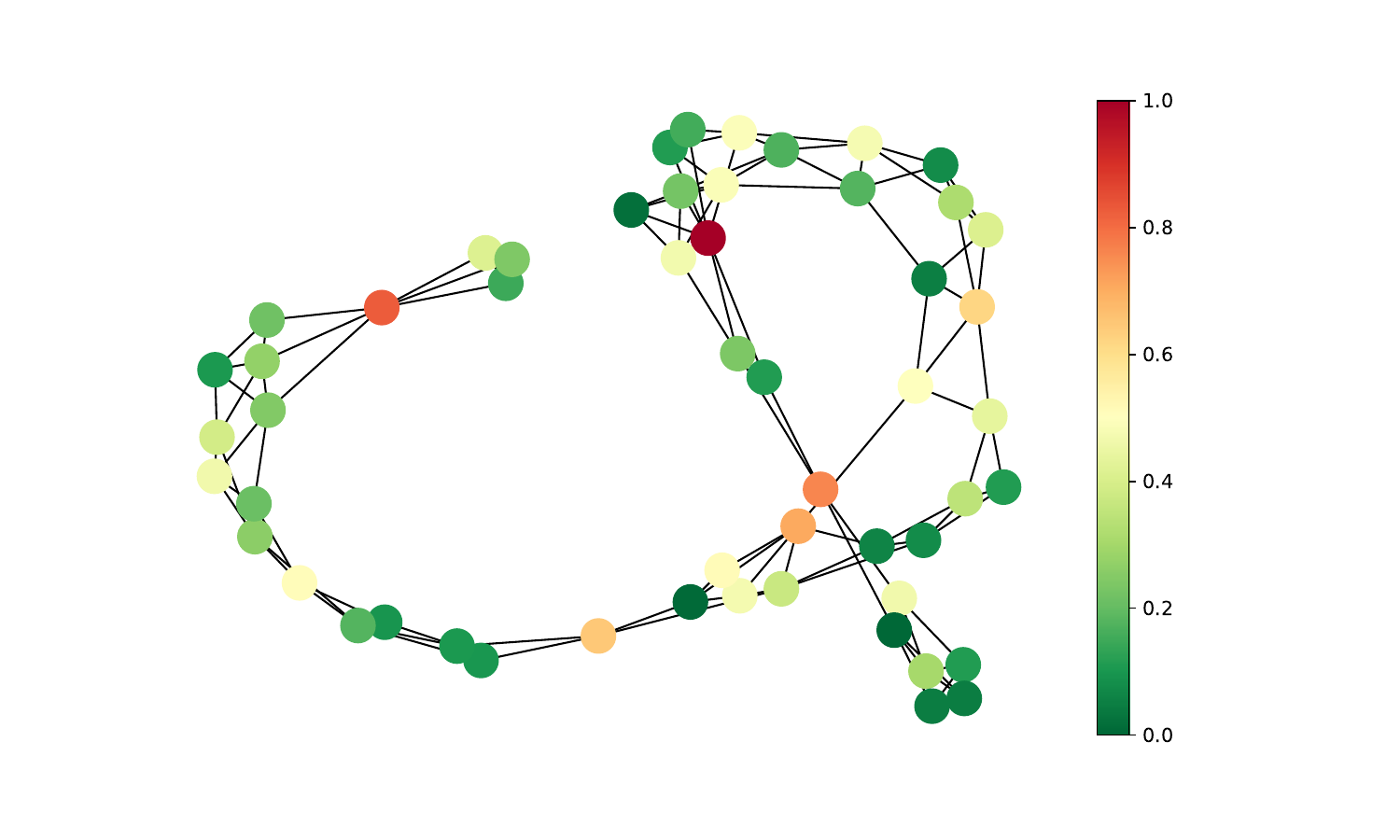}}
    \subfigure[8 models]{\centering \includegraphics[trim=0cm 2cm 5cm 2cm, clip,width=.2\linewidth]{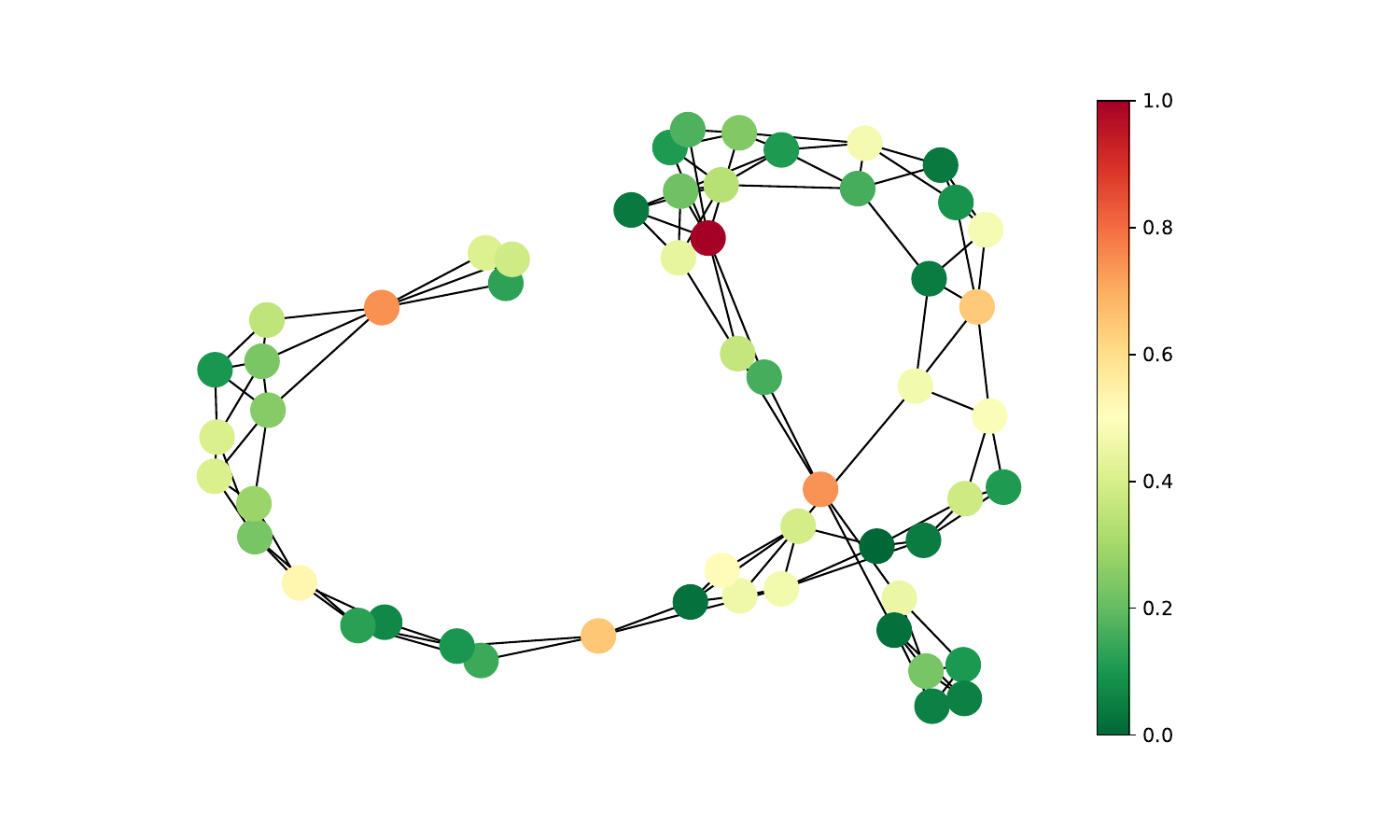}}

    \vspace{-0.2cm}
    \caption{\small (a) 3D structure of protein Adenylosuccinate lyase, (b)$\sim$(e) Pruning score of nodes learned with MVP attached mincut pool in various number of view, (f) Important nodes in Adenylosuccinate lyase, (g) Attention score of nodes learned with SAGpool, (h)$\sim$(j) Pruning score of nodes in ensemble model with various number of mincut pool models}
    \vspace*{-0.7cm}
    \label{fig:3d_image}
\end{figure}

\vspace*{-0.2cm}
\paragraph{Qualitative analysis.}
We further analyze examples from the PROTEINS data to qualitatively examine the effect of our graph pruning method. Figure~\ref{fig:3d_image} shows the 3D structure and pruned graphs from the PROTEINS dataset, in which the nodes with higher pruning scores are colored as red and nodes with lower scores are colored green. 
Specifically, Figure~\ref{fig:3d_image}(a) shows the 3D structure of adenylosuccinate lyase, which catalyzes the reactions that convert adenylosuccinate to AMP in the purine biosynthetic pathway. 
In enzymes, the binding site is the most important node since it directly binds with the substrate to cause enzyme reactions. 
The pruning scores of the nodes that correspond to the binding site are high with attention-based scoring and ensemble models as shown in Figure~\ref{fig:3d_image}(g)$\sim$ (j), which results in the pruning of these critical nodes for enzyme classification. However, with MVP scoring, their pruning scores are computed low. MVP sets the pruning score high for nodes that are less important in representing the protein, and are unmatched with the binding site.
We also see that the pruned parts in the 3D structure do not affect the prediction of whether the given protein structure is an enzyme or not. 
This shows evidence that our model prunes the graph without losing the most crucial information about the given graph for a target task, that agrees with the domain knowledge.


\begin{table}[bt]
    \centering
    \captionof{table}{\small Loss Ablation on multiple losses. We compare the performance of the models based on $L_{CE}$, then add $L_{pool}$ or $L_r$. $L_{CE},L_{pool},L_r$ are cross entropy, pool, and reconstruction loss each}
        \resizebox{\textwidth}{!}{
        \begin{tabular}{l|c c c c|c c c c}
    \thickhline 
    Dataset & \multicolumn{4}{c}{PROTEIN} & \multicolumn{4}{c}{DD}\\\hline
    Method & Original & L$_{CE}$& with L$_{pool}$ & with L$_r$ &  Original & L$_{CE}$ & with L$_{pool}$ & with L$_r$\\\hline
    MVP+GMT & \textbf{79.46$_{\pm 2.22}$}
            & 75.98$_{\pm 4.10}$
            & -
            & -
            & \textbf{77.46$_{\pm 2.15}$}
            & 75.17$_{\pm 3.15}$
            & -
            & - \\
    MVP+MinCut & \textbf{81.70$_{\pm 3.10}$}
            & 77.86$_{\pm 4.57}$ 
            & 79.20$_{\pm 4.09}$ 
            & 79.46$_{\pm 3.12}$ 
            & \textbf{82.88$_{\pm 2.42}$}
            & 79.75$_{\pm 3.60}$ 
            & 78.31$_{\pm 4.29}$ 
            & 80.59$_{\pm 4.02}$  \\
    MVP+Diff & \textbf{79.91$_{\pm 4.38}$} 
            & 76.70$_{\pm 1.97}$ 
            & 76.61$_{\pm 4.98}$ 
            & 78.39$_{\pm 3.95}$ 
            &\textbf{76.16$_{\pm 3.04}$}
            & 79.41$_{\pm 3.17}$ 
            & 75.76$_{\pm 5.69}$ 
            & 80.34$_{\pm 2.00}$ \\
    \thickhline
    \end{tabular}
    }
    \label{tab:loss_abl}
    \vspace{-0.3cm}
\end{table}

\vspace*{-0.3cm}

\section{Conclusion}
\vspace*{-0.3cm}
We show that existing attention-based node pruning methods result in simple pruning of nodes based on their degrees, since they will be assigned higher attention scores as they capture information about more nodes. Since this can result in removing critical nodes for the given task, we proposed a novel graph pooling method that can more carefully set the score for each node in balanced perspectives, by utilizing multi-view embeddings and reconstruction errors. Specifically, we construct a multi-view embedding either from intrinsic modalities of the data, or artificially constructed views from a random projection of the subset of the features. Then, we attempt to reconstruct the original nodes and the adjacency matrix using the concatenated multi-view features, and detect the nodes with high reconstruction errors as less informative ones. Using such multi-view reconstruction-based pruning scores, we then prune the nodes whose scores are higher than a set threshold. Our method can be combined with any existing graph pooling methods, to obtain a more compact, task-relevant subgraph of the given graph. We validate our method, MVP, on multiple real-world benchmark datasets, to show that model outperforms state-of-the-art graph pooling methods. We further show that this performance improvement comes from its consideration of rich node features, as its pruning score is not dominated by the topological features, unlike the pruning scores of existing methods. We further perform an in-depth analysis of our model to examine where the performance improvements come from, and whether our method does prune nodes based on their importance, according to the domain knowledge. We describe the \textbf{limitations and societal impacts} in the supplementary file, due to the space limitation. 

\newpage
\bibliography{reference}
\bibliographystyle{nips}


\clearpage
\appendix
\label{supple}
\textbf{Supplementary Material}
\normalsize
\section{Analysis of Multi-view Embedding}\label{supple:tsne}

\begin{figure}[hbt!]
    \centering
    \includegraphics[width=.8\linewidth]{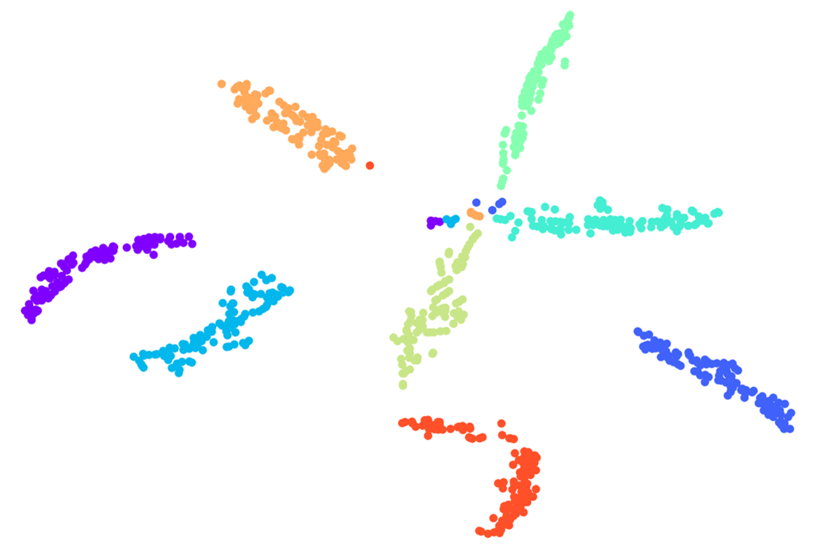}
    \caption{t-SNE visualization of the embedded features for 8 randomly constructed views. Each color denotes a different view.}
    \label{fig:tSNE-multi_view}
\end{figure}

We split the input features into 8 views and embed them separately. Figure \ref{fig:tSNE-multi_view} shows the visualization of the t-SNE embeddings for each view.
As shown, the divided features are clustered into separate and individual clusters. This suggests that randomized projection of the input features captures the multi-modality of the data well. 

\begin{figure}[hbt!]
    \centering
    \subfigure[\small Important nodes]{\includegraphics[height=.2\linewidth,trim=3.5cm 2cm 6.5cm 2cm, clip]{FIG/critical_fig.pdf}}
    \hspace{1em}
    \subfigure[\small model 1]{\includegraphics[height=.2\linewidth,trim=3.5cm 2cm 6.5cm 2cm, clip]{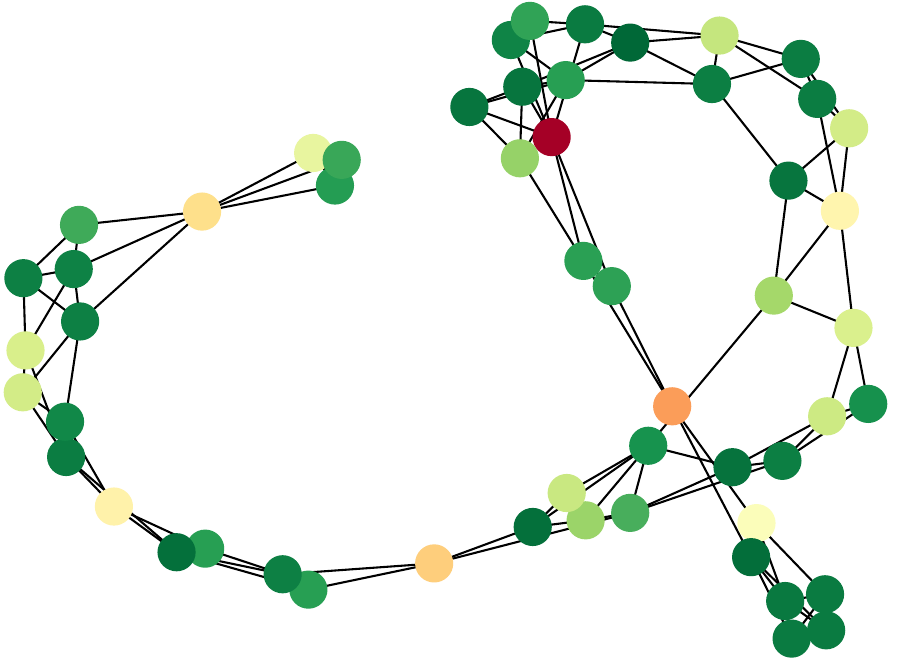}}
    \subfigure[\small model 2]{\includegraphics[height=.2\linewidth,trim=3.5cm 2cm 4.85cm 2cm, clip]{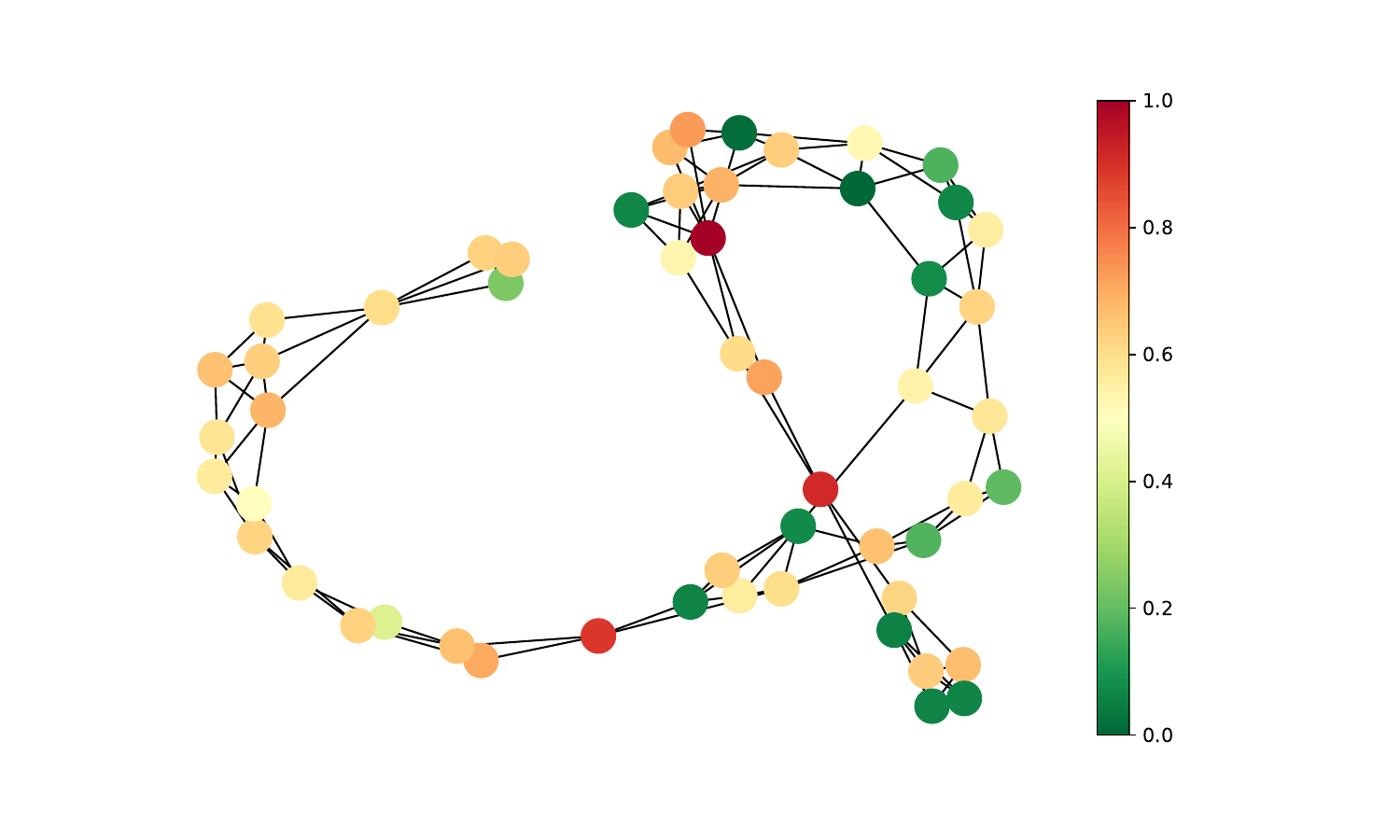}}
    \caption{Pruning score of nodes in each model of ensemble with 2 models}
    \label{fig:2ensemble_supple}
\end{figure}

\begin{figure}[hbt]
    \centering
    \subfigure[\small model 1]{\includegraphics[height=.16\linewidth,trim=3.5cm 2cm 6.5cm 2cm, clip]{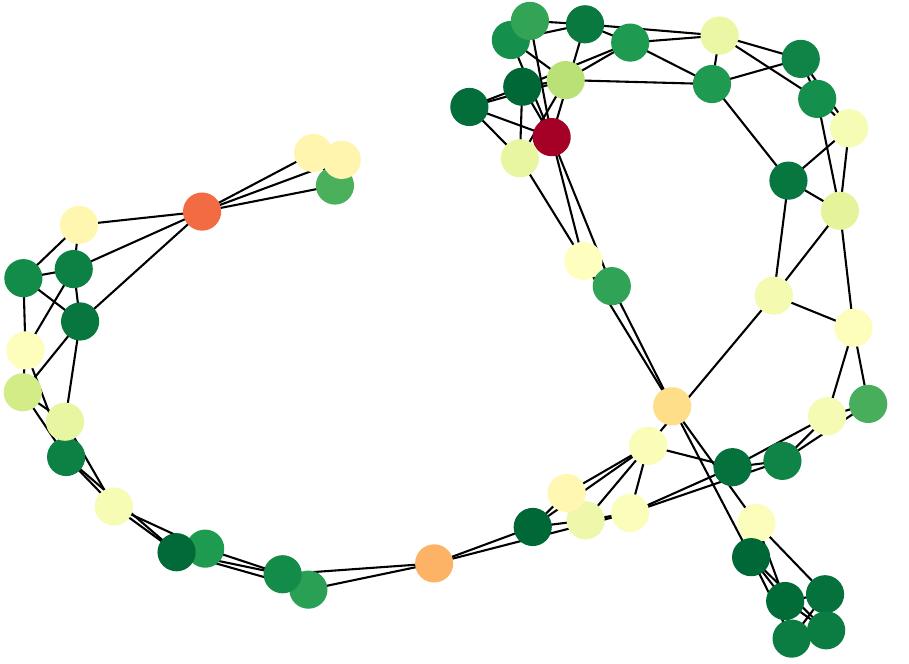}}
    \subfigure[\small model 2]{\includegraphics[height=.16\linewidth,trim=3.5cm 2cm 6.5cm 2cm, clip]{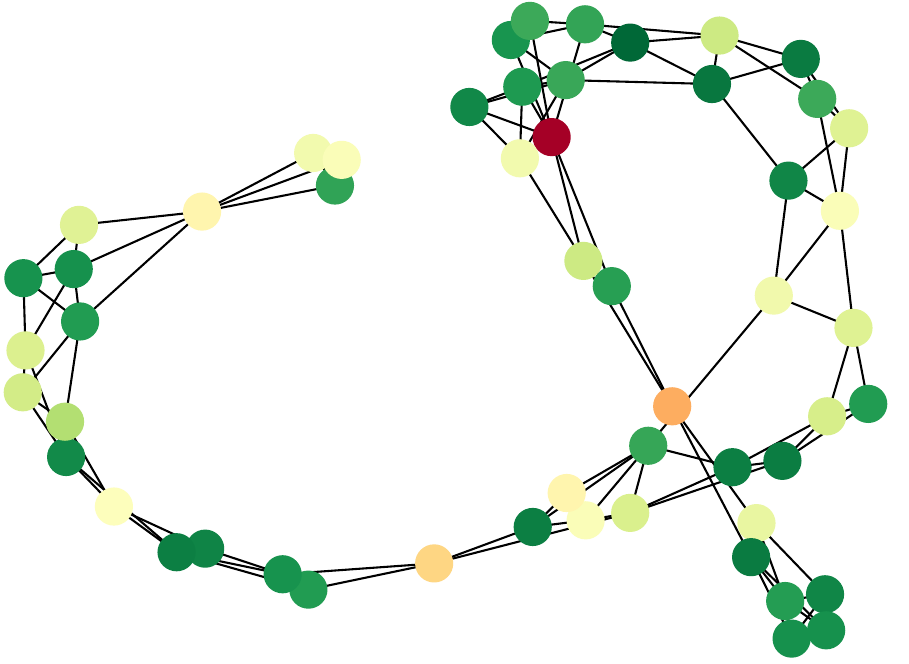}}
    \subfigure[\small model 3]{\includegraphics[height=.16\linewidth,trim=3.5cm 2cm 6.5cm 2cm, clip]{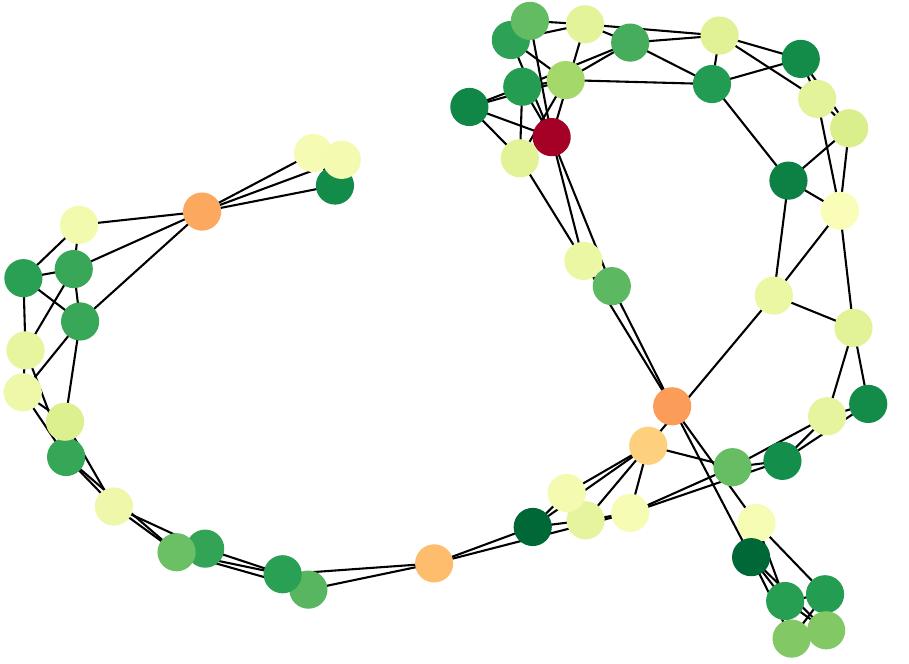}}
    \subfigure[\small model 4]{\includegraphics[height=.16\linewidth,trim=3.5cm 2cm 4.85cm 2cm, clip]{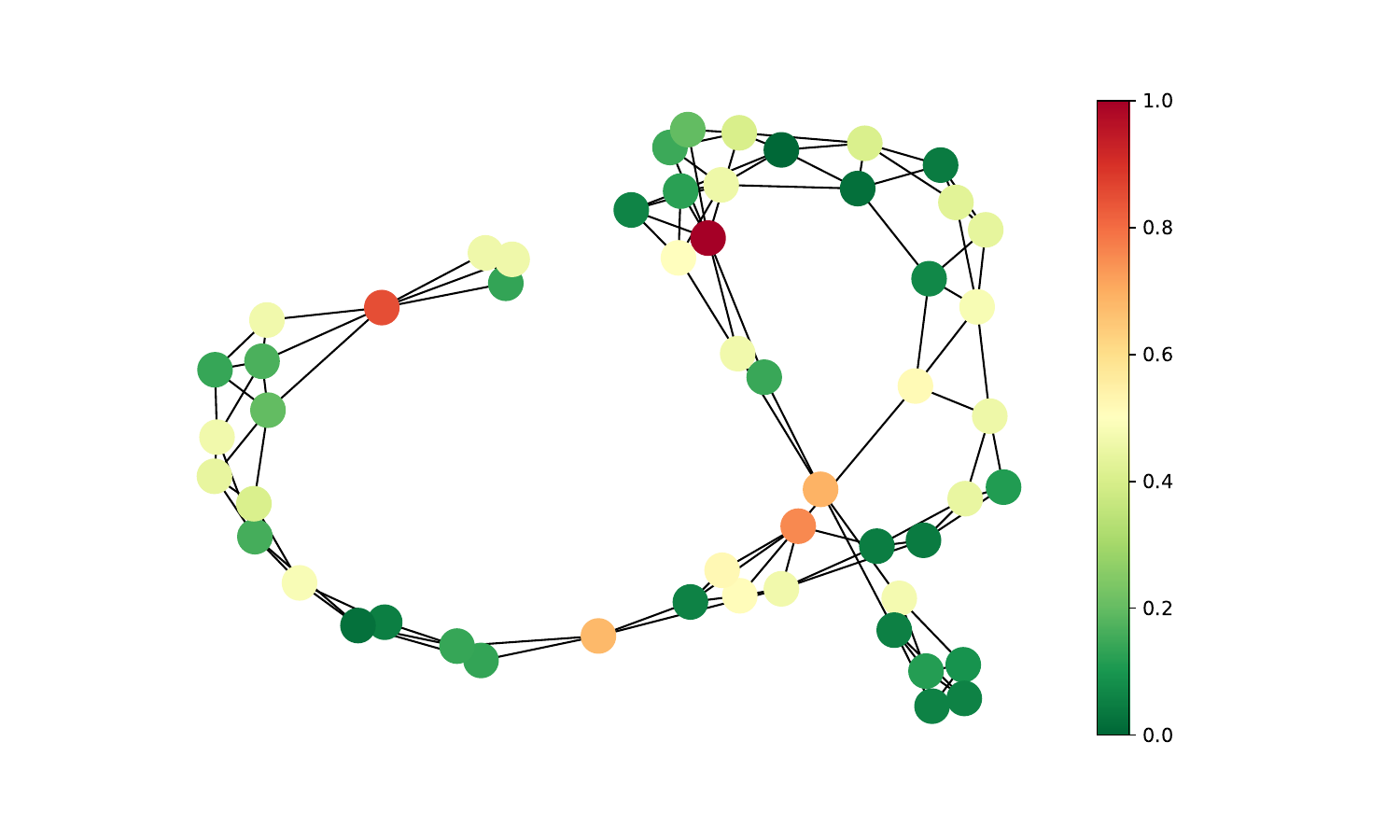}}
    \caption{Pruning score of nodes in each model of ensemble with 4 models}
    \label{fig:4ensemble_supple}
\end{figure}

\begin{figure}[hbt]
    \centering
    \subfigure[\small model 1]{\includegraphics[height=.16\linewidth,trim=3.5cm 2cm 6.5cm 2cm, clip]{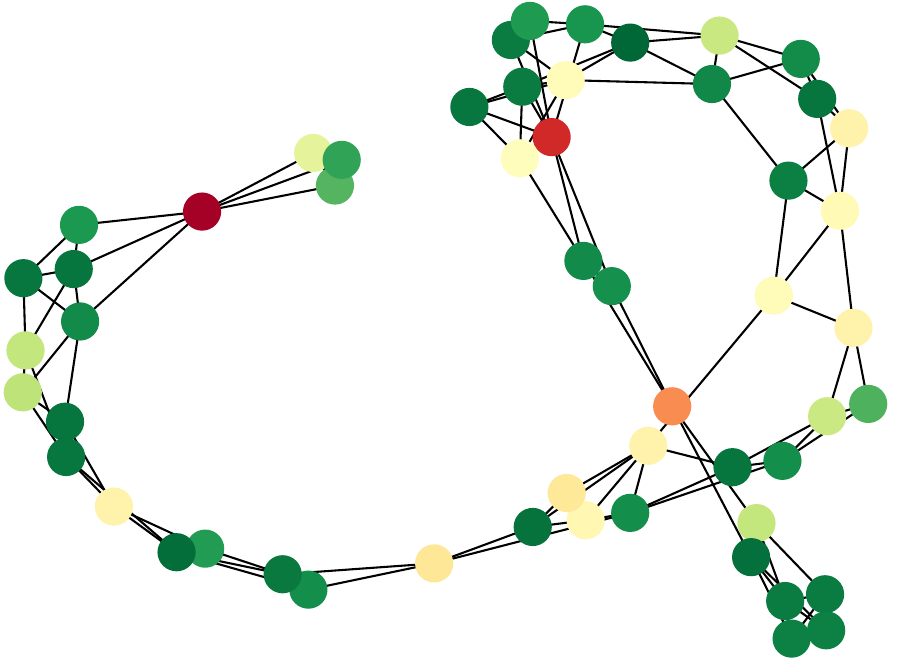}}
    \subfigure[\small model 2]{\includegraphics[height=.16\linewidth,trim=3.5cm 2cm 6.5cm 2cm, clip]{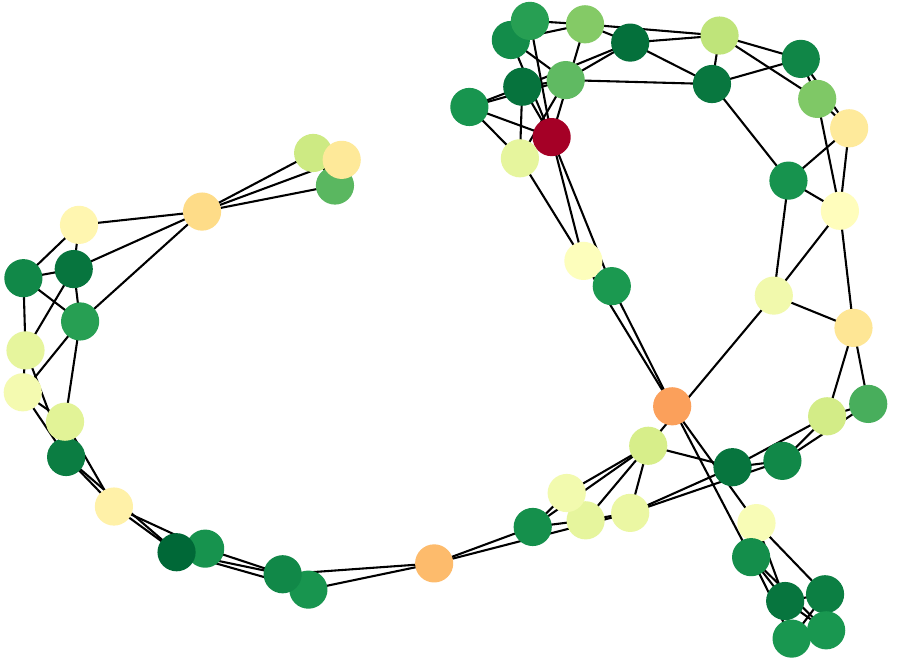}}
    \subfigure[\small model 3]{\includegraphics[height=.16\linewidth,trim=3.5cm 2cm 6.5cm 2cm, clip]{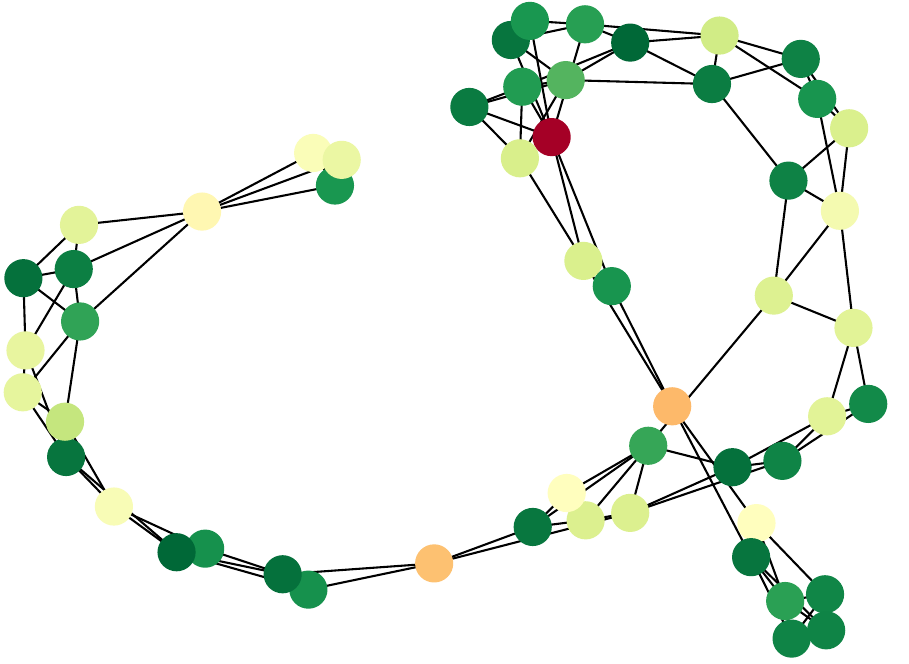}}
    \subfigure[\small model 4]{\includegraphics[height=.16\linewidth,trim=3.5cm 2cm 6.5cm 2cm, clip]{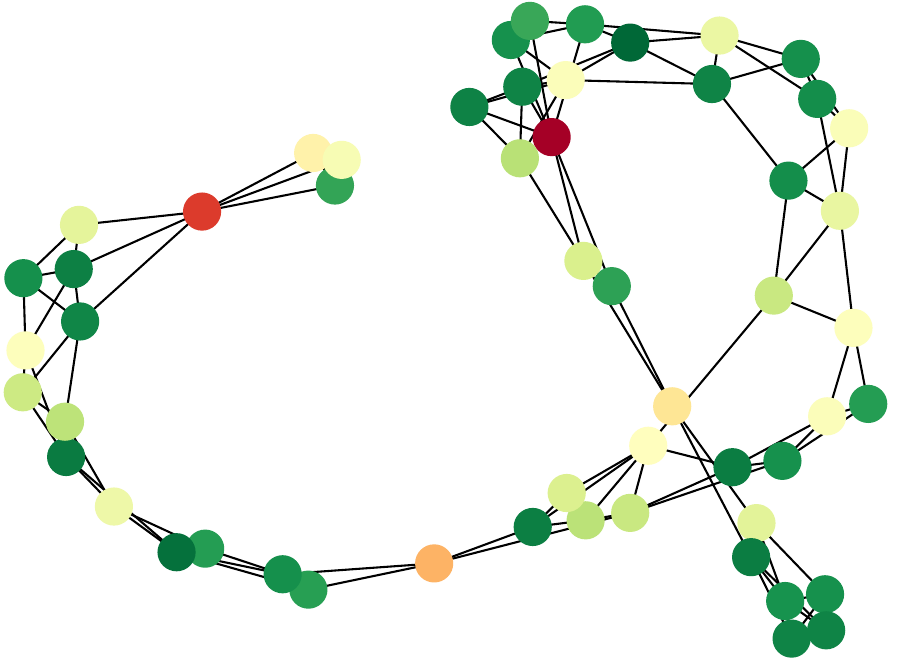}}
    
    \subfigure[\small model 5]{\includegraphics[height=.16\linewidth,trim=3.5cm 2cm 6.5cm 2cm, clip]{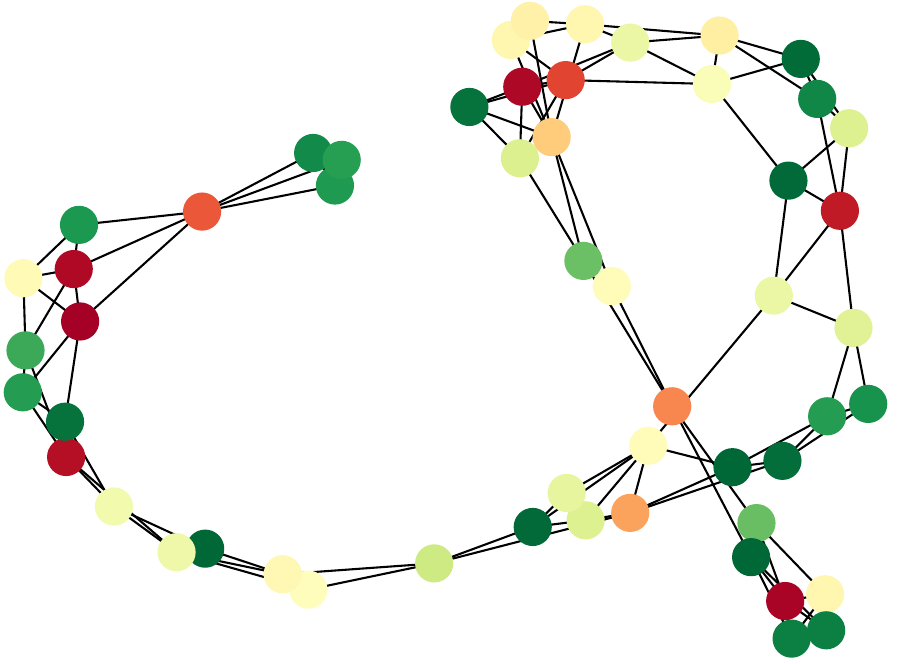}}
    \subfigure[\small model 6]{\includegraphics[height=.16\linewidth,trim=3.5cm 2cm 6.5cm 2cm, clip]{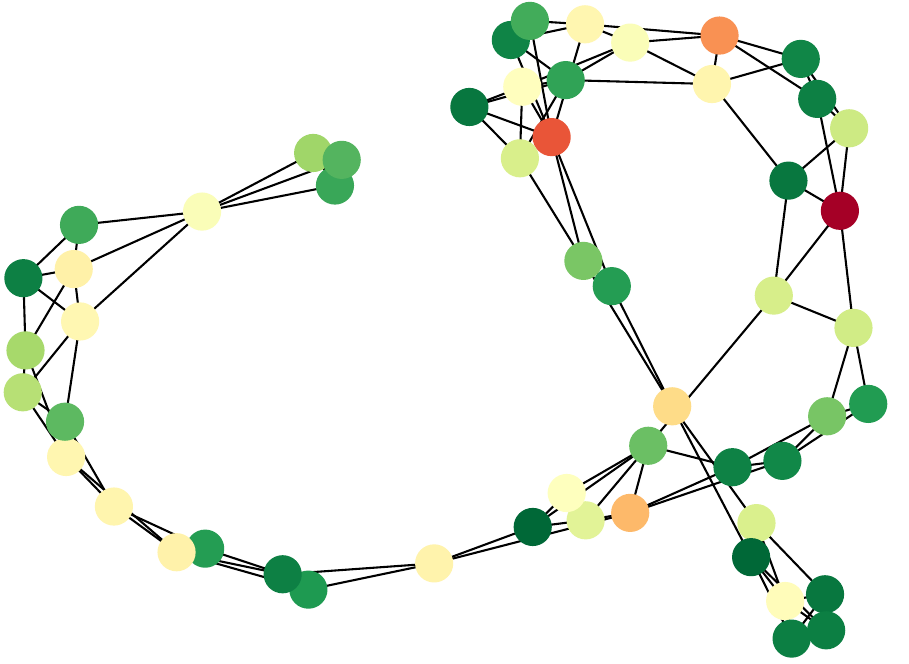}}
    \subfigure[\small model 7]{\includegraphics[height=.16\linewidth,trim=3.5cm 2cm 6.5cm 2cm, clip]{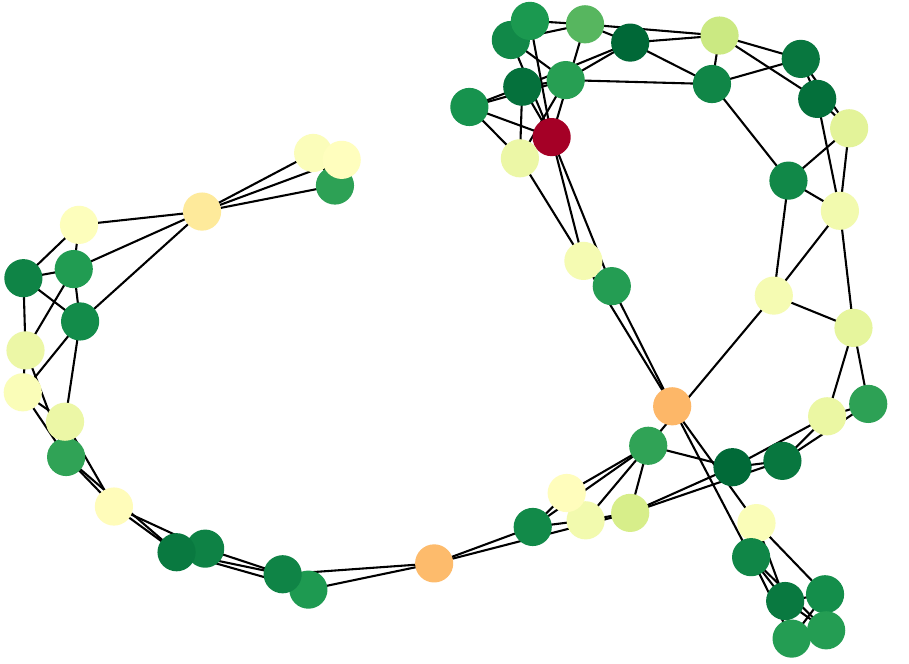}}
    \subfigure[\small model 8]{\includegraphics[height=.16\linewidth,trim=3.5cm 2cm 4.85cm 2cm, clip]{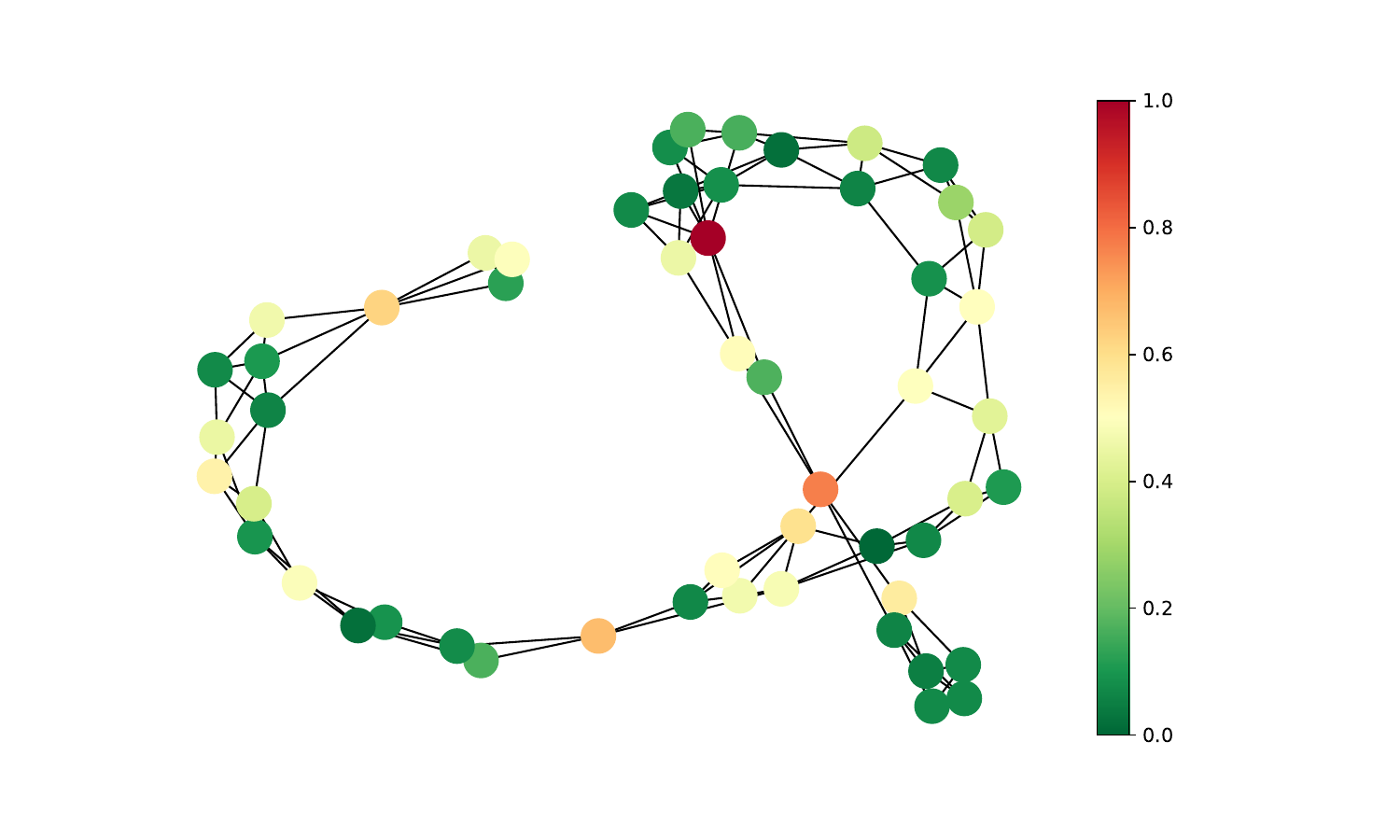}}
    \caption{Pruning score of nodes in each model of ensemble with 8 models}
    \label{fig:8ensemble_supple}
\end{figure}

\textcolor{blue}{Figure \ref{fig:2ensemble_supple}, \ref{fig:4ensemble_supple}, and \ref{fig:8ensemble_supple} show the pruning scores of nodes in each model of the ensemble, which are shown using the average as representative values in main paper Figure \ref{fig:3d_image}. 
As shown in Figure \ref{fig:2ensemble_supple}, \ref{fig:4ensemble_supple}, and \ref{fig:8ensemble_supple}, the most of the binding sites are not considered as important nodes in ensemble method. Also, this tendency is shown in all of the ensemble models no matter how many models are learned in the ensemble method.
These results indicates that MVP is not effect same as the ensemble method, while they are novel model considering important part of the graph based on multi-view framework.}

\section{Datasets}\label{supple:dataset}
We test our model and baselines on various benchmark datasets (TU dataset\footnote{https://ls11-www.cs.tu-dortmund.de/staff/morris/graphkerneldatasets}, OGB dataset\footnote{https://ogb.stanford.edu/docs/graphprop/}) for graph classification.
PROTEINS~\cite{borgwardt2005protein, dobson2003distinguishing} and DD~\cite{dobson2003distinguishing,shervashidze2011weisfeiler} are datasets containing protein tertiary structures, where each protein is represented by a graph, and the nodes are secondary structure elements of protein. The goal of the task is to classify whether the given protein is an enzyme or not.
\textcolor{blue}{HIV~\cite{hu2021open} is a molecular property prediction datasets.}
FRANKENSTEIN~\cite{orsini2015graph} is a modified version of the BURSI dataset~\cite{kazius2005derivation}, which discards the Bond-type information and remaps the most frequent atom symbols (vertex labels) to MNIST digit images. The original atom symbols can only be recovered through the high dimensional MNIST vectors of pixel intensities.
NCI1~\cite{wale2008comparison} contains two balanced subsets of the dataset of chemical compounds screened for activity against non-small cell lung cancer.
COLLAB~\cite{yanardag2015deep} is a scientific collaboration dataset composed of three public datasets~\cite{leskovec2005graphs}, namely, High Energy Physics, Condensed Matter Physics, and Astro Physics.  
IMDB~\cite{rossi2015network} is a dataset constructed from the actors/actresses and the genre information of different movies on the IMDB website. Table~\ref{tab:acc} contains data statistics for all datasets used for our main experiment. Raw data is taken from the repository of benchmark dataset for graph kernels\footnote{ls11-www.cs.tu-dortmund.de/staff/morris/graphkerneldatasets}. In all experiments, the PROTEINS dataset is mainly used.

\section{parameter overhead}\label{supple:param}

\begin{figure}[hbt!]
    \centering
    \includegraphics[width=.7\linewidth]{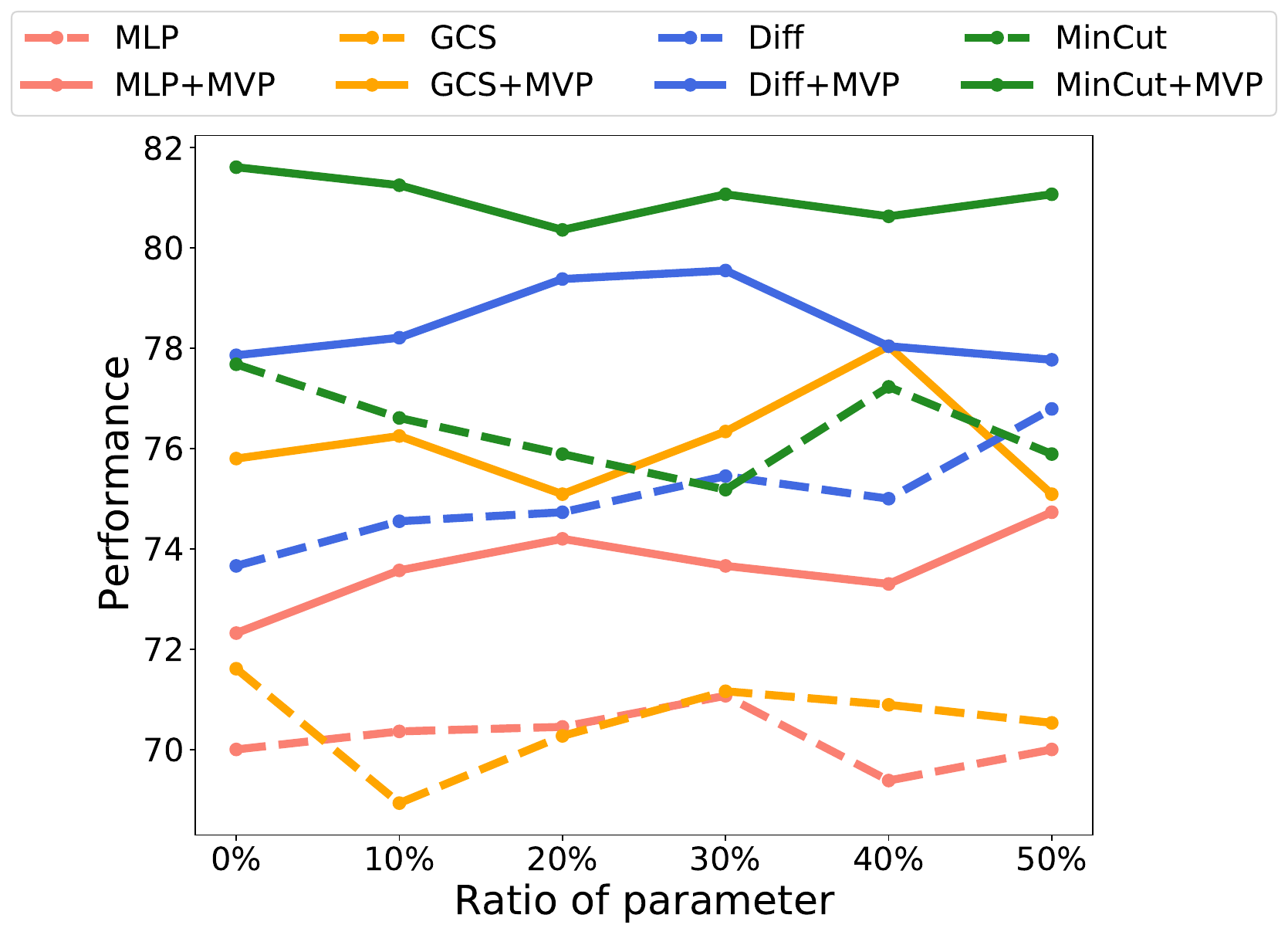}
    \caption{The performance for each parameter ratio of PROTEIN dataset.}
    \label{fig:param_test_supple}
\end{figure}

\definecolor{lavendermist}{rgb}{0.95, 0.95, 1.0}
\begin{table}[hbt!]
    \centering
    \caption{Parameter test on original pooling methods and methods with MVP layer using PROTEIN dataset. We compared the accuracy by increasing the parameter by 10\% based on the parameter of the original method, and increasing the ratio until it became 1.5.}
    \resizebox{\textwidth}{!}{\begin{tabular}{l c c c c c c}
        \thickhline
        \multirow{2}{*}{Method} & \multicolumn{6}{c}{ratio (test parameter/original parameter)}
        \\\cmidrule(lr){2-7}
         & 1 & 1.1 & 1.2 & 1.3 & 1.4 & 1.5
         \\\hline
         \\[-1em]
        MLP & 70.00$_{\pm 4.23}$ & 70.36$_{\pm 4.37}$ & 70.45$_{\pm 2.20}$ & 71.07$_{\pm 2.74}$ & 69.38$_{\pm 1.79}$ & 70.00$_{\pm 4.30}$
        \\
        \rowcolor{lavendermist}
        MLP+MVP & 72.32$_{\pm 1.96}$ & 73.57$_{\pm 70.36}$ & 74.20$_{\pm 3.54}$ & 73.66$_{\pm 3.96}$ & 73.30$_{\pm 3.10}$ & 74.73$_{\pm 2.96}$
        \\
        GCS & 71.61$_{\pm 2.87}$ & 68.93$_{\pm 2.52}$ & 70.27$_{\pm 4.11}$ & 71.16$_{\pm 3.22}$ & 70.89$_{\pm 2.33}$ & 70.53$_{\pm 3.55}$
        \\
        \rowcolor{lavendermist}
        GCS+MVP & 75.80$_{\pm 4.27}$ & 76.25$_{\pm 5.18}$ & 75.09$_{\pm 3.25}$ & 76.34$_{\pm 3.73}$ & 78.04$_{\pm 4.97}$ & 75.09$_{\pm 2.72}$
        \\
        Diffpool & 73.66$_{\pm 4.68}$ & 74.55$_{\pm 3.88}$ & 74.73$_{\pm 3.57}$ & 75.45$_{\pm 4.48}$ & 75.00$_{\pm 3.87}$ & 76.79$_{\pm 4.74}$
        \\
        \rowcolor{lavendermist}
        Diffpool+MVP & 77.86$_{\pm 3.54}$ & 78.21$_{\pm 2.83}$ & 79.38$_{\pm 3.94}$ & 79.55$_{\pm 3.16}$ & 78.04$_{\pm 4.11}$ & 77.77$_{\pm 3.33}$
        \\
        Mincut & 77.77$_{\pm 2.54}$ & 76.61$_{\pm 2.45}$ & 75.89$_{\pm 3.39}$ & 75.18$_{\pm 4.39}$ & 77.23$_{\pm 2.56}$ & 75.89$_{\pm 2.88}$
        \\
        \rowcolor{lavendermist}
        Mincut+MVP & 81.61$_{\pm 3.15}$ & 81.25$_{\pm 4.09}$ & 80.36$_{\pm 1.55}$ & 81.07$_{\pm 4.90}$ & 80.63$_{\pm 4.15}$ & 81.07$_{\pm 2.70}$
        \\\thickhline
    \end{tabular}
    }
    \label{tab:param_test}
\end{table}

\textcolor{blue}{We verify that MVP does not simply obtain improved performance by using more parameters. We compared the accuracy of MVP and other baselines except GMT, because GMT has about ten times more parameters than our method MVP. As shown in Figure \ref{fig:param_test_supple} and Table \ref{tab:param_test}, the increase of the parameter shows a marginal improvement, but the performance gap between original baseline and method with MVP shows significantly large gap.}

\section{
Limitation and Societal impacts}\label{supple:societal}

\textcolor{blue}{Our study contains limitations of conventional machine learning. Since our work utilizes the usual training procedure of graph neural networks, it may reflect potential biases such as data collection bias which can be presented in supervised training methods. The development and success of various graph neural networks have potential to be used to solve practical problems. Our model is also a work that improves the development of these graph neural networks, so we have to keep in mind the general ethical issues that may arise in domains such as social and biochemical domains where network knowledge is used in various ways.}

\end{document}